# Predicting Consultation Success in Online Health Platforms Using Dynamic Knowledge Networks and Multimodal Data Fusion

Shuang Geng, Ph.D.
Shenzhen University
1066 Xueyuan Road
Shenzhen, China 518055
Email: gs@szu.edu.cn

Wenli Zhang, Ph.D.
Iowa State University
3332 Gerdin, 2167 Union Drive
Ames, IA, USA 50011-2027
Email: wlzhang@iastate.edu

Jiaheng Xie, Ph.D.
University of Delaware
217 Purnell Hall
Newark, DE, USA 19716
Email: jxie@udel.edu

Gemin Liang
Shenzhen University
1066 Xueyuan Road
Shenzhen, China 518055
Email: 2100132025@email.szu.edu.cn

Ben Niu
Shenzhen University
1066 Xueyuan Road
Shenzhen, China 518055
Email: drniuben@gmail.com

Sudha Ram
University of Arizona
McClelland Hall 430J, 1130 E. Helen St.
Tucson, Arizona 85721-0108
Email: ram@eller.arizona.edu

Please send comments to Shuang Geng at gs@szu.edu.cn or Wenli Zhang at wlzhang@iastate.edu.

# Predicting Consultation Success in Online Health Platforms Using Dynamic Knowledge Networks and Multimodal Data Fusion

## ABSTRACT

Online healthcare consultation in virtual health is an emerging industry marked by innovation and fierce competition. Accurate and timely prediction of healthcare consultation success can proactively help online platforms address patient concerns and improve retention rates. However, predicting online consultation success is challenging due to the partial role of virtual consultations in patients' overall healthcare journey and the disconnect between online and in-person healthcare IT systems. Patient data in online consultations is often sparse and incomplete, presenting significant technical challenges and a research gap. To address these issues, we propose the **Dy**namic **Kno**wledge **Ne**twork and **M**ultimodal Data Fusion (DyKoNeM) framework, which enhances the predictive power of online healthcare consultations. Our work has important implications for new business models where specific and detailed online communication processes are stored in the IT database, and at the same time, latent information with predictive power is embedded in the network formed by stakeholders' digital traces. It can be extended to diverse industries and domains, where the virtual or hybrid model (e.g., integration of online and offline services) is emerging as a prevailing trend.

## INTRODUCTION

Online healthcare consultation belongs to virtual health – an emerging industry that integrates public health, medical informatics, and healthcare businesses (Eysenbach, 2001; HHS, 2023). Virtual health encompasses telehealth (e.g., synchronous telemedicine, asynchronous provider-to-patient consultation, and remote patient monitoring), digital therapeutics (e.g., replacement therapies and treatment optimization), and care navigation (e.g., patient self-directed care and e-triage) (Fowkes et al., 2020). The most significant advantage of virtual health lies in its ability to overcome the challenge of traditional health services that are contingent on the physical presence of doctors and patients at the same time and location (Wootton et al., 2017). This opens up a new venue for innovative healthcare delivery, benefiting not only developing countries but also industrialized countries with healthcare disparities.

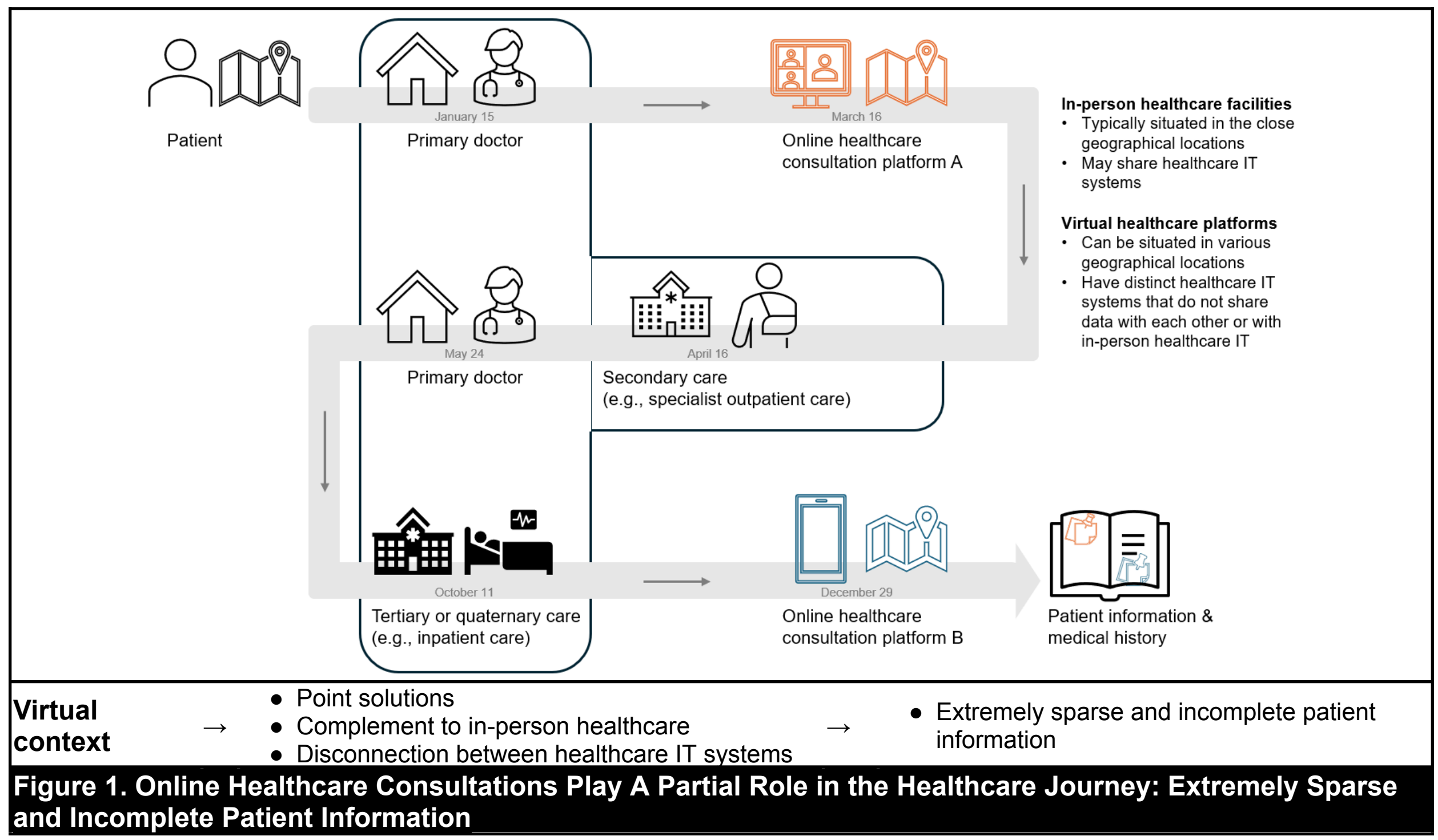

**Figure 1. Online Healthcare Consultations Play A Partial Role in the Healthcare Journey: Extremely Sparse and Incomplete Patient Information**

Compared to in-person healthcare, in the virtual context, consultation success (i.e., the resolution of patients' health questions or problems in a way that the patient is satisfied; a formal definition can be found in Section: Problem Formulation) and patient retention emerge as particularly crucial considerations due to intense competition. In in-person healthcare, patients often select their healthcare services from a restricted pool due to factors such as geographical distance, insurance restrictions, and adherence to their primary physicians. Virtual health platforms have removed geographic and time barriers of in-person healthcare, where doctors and patients are required to be present in the same location at the same time. This is a double-edged sword for patient retention in the virtual context: on the one hand, it has the potential to draw patients from diverse regions, allowing them to engage in healthcare consultations at their convenience; on the other hand, it also implies that the online healthcare platforms need to compete with in-person health services in the patients' respective locations, as well as with other online platforms (Y. Li et al., 2019). In fact, for online healthcare consultations, patients' dissatisfaction can significantly impact their retention (Fowkes et al., 2020). Naturally, improving consultation success rates has become a strategic goal for many online healthcare platforms and is crucial for maintaining a competitive edge in this fiercely competitive industry (Fowkes et al., 2020). By accurately and promptly predicting consultation success, online platforms can take proactive steps to address patients' concerns and needs, potentially improving patient retention rates. For instance, leveraging the prediction results, the platform can send reminders to healthcare providers, prompting them to focus more on particular aspects such as efficacy, interaction manner, or empathy. Decision support systems could also be devised, utilizing the prediction outcomes to assist doctors in streamlining the consultation process – for example, offering patient education materials outlining correct expectations for consultations, thereby reducing the rate of failed consultations, as well as providing aftercare education. Additionally, if a patient is predicted to have a high likelihood of experiencing a failed consultation (e.g., not receiving an effective response from the doctor within the first few interactions, the doctor failing to provide explicit operational instructions, or not showing sufficient empathy or emotional support), the platform could recommend alternative doctors before losing the patient.

Compared to in-person healthcare, predicting the success of online healthcare consultations poses two distinct challenges, visualized in Figure 1. First, the patient data presented on online healthcare consultation platforms often appear fragmented. This is because, at the current time, online health consultation (or virtual health in general) is generally perceived as a supplementary method to in-person healthcare. For example, it provides quick access to medical advice when in-person consultations are inconvenient or when seeking a second opinion in cases of significant illnesses. Meanwhile, there is a widespread prevalence of point solutions in virtual health, which involve the deployment of specific, focused IT solutions addressing particular aspects of patient needs. For instance, customized online chat software may be employed for text-based patient-doctor communication. Consequently, online healthcare consultation plays a partial role in the overall patient healthcare journey. Second, there is often a disconnect between online health consultation and in-person healthcare IT systems. In a given geographical region or within the same group or company, in-person healthcare facilities may collaborate through a shared suite of healthcare IT systems, thereby enabling the exchange of medical data and patient medical histories. In contrast, online health consultations constitute a more recent development associated with emerging healthcare companies. These platforms typically operate on separate healthcare IT systems, functioning independently of in-person healthcare IT systems, due to geographic limitations, legal and privacy regulations, and potential conflicts of interest with established healthcare practices. Consequently, data integration between online platforms and in-person healthcare IT systems is typically not realized. Due to these two characteristics, patient data presented within the online healthcare consultation context tends to be notably sparse and incomplete. There are significant challenges in acquiring a comprehensive medical history or demographic information from patients, and obtaining details about a patient's disease progression becomes exceedingly challenging. Furthermore, the inherently "virtual" nature of online healthcare consultation complicates the process of accessing a patient's auxiliary test results, such as laboratory and imaging tests. Ultimately, in a novel context of online healthcare consultation, the confluence of data sparsity and challenges in acquiring comprehensive patient information imply significant technical challenges and highlight a notable research gap.

To address the research gaps in predicting healthcare consultation success in the virtual context and fully leverage the data advantages not present in traditional in-person settings, guided by the design science paradigm (Gregor & Hevner, 2013), we devise a **Dy**namic **Kno**wledge **Ne**twork and **M**ultimodal Data Fusion (DyKoNeM) framework. DyKoNeM leverages state-of-the-art language models with temporal information to represent *explicit knowledge*, namely patient-doctor communication processes to discern patterns of interaction. Furthermore, informed by Homophily theory, we introduce a novel Knowledge Graph Attention Network tailored to capture nuanced and evolving *implicit knowledge*; specifically, multi-view (online and offline) dynamic networks that have formed through digital traces of patients' interactions with various stakeholders during their healthcare journey. Additionally, to effectively fuse the *explicit* and *implicit knowledge*, we propose a new multi-modal data fusion method that addresses the distinct data distribution from different data modalities while ensuring efficiency. Consequently, DyKoNeM greatly enhances the predictive power of healthcare consultation success in online healthcare platforms.

Our contributions are twofold. From the domain perspective, we present a novel framework for predicting patient consultation success in the virtual context. Collaborating with one of China's largest virtual health platforms and employing online healthcare consultations as a research case, we demonstrate that our method outperforms existing healthcare consultation success prediction methods significantly. Our framework offers an effective solution for online healthcare platforms to predict the success of an individual consultation transaction. Subsequently, platforms can take proactive measures to cater to the specific needs and concerns of patients, which can increase the likelihood of retaining patients and ultimately enhance healthcare quality and patient outcomes. From the design science perspective, our technical contributions are as follows: (1) Within the devised framework, we introduce a new Homophily theory-based knowledge network that distinguishes itself from existing work by its capacity to represent: a) multi-view perspectives; b) nodes (e.g., stakeholders) attributes encoding and propagation; c) network dynamics. (2) As part of DyKoNeM, our multimodal data fusion differs from existing approaches by a) considering the distinct data distributions inherent in different data modalities; b) striving for a balance between data fusion efficiency and comprehensive representation. These contributions collectively enhance the predictive capabilities for online healthcare consultation, while also advancing the methodology in design science.

Our work has important implications for IS research. First, we contribute to healthcare predictive analysis (Baird et al., 2018). While our work focuses on overcoming the challenges associated with virtual health, an emerging industry, specifically online healthcare consultation, the phenomenon we underscore is widespread in today's digitized healthcare industry: on the one hand, there is specific and detailed *explicit knowledge* (e.g., detailed online communication processes) in the IT database, and on the other hand, *implicit knowledge* is formed by stakeholders' digital traces. Second, as digital transformation continues to infiltrate various industries, the method we proposed demonstrates the effective integration of explicit knowledge alongside implicit knowledge, which possesses the potential to catalyze the advancement of predictive analytics in diverse industries and domains, where the virtual or hybrid model (e.g., integration of online and offline services) is emerging as a prevailing trend.

## RESEARCH CONTEXT AND DATA

### Business Partner and Dataset Summary

We have partnered with one of the leading virtual health platforms in China, JianKang160. JianKang160 was founded in 2005 with the support and close cooperation of the local government. Since its foundation, it has gradually expanded its cloud-based solutions for hospitals and virtual health services for patients across China. It now provides a variety of services, such as setting up appointments for offline hospital visits, online consultations, online physical exams, and online pharmacies. To date, JianKang160 has forged collaborations with more than 25,000 hospitals and organizations, positioning it as the largest virtual health platform in China based on partner hospital magnitude. Its virtual health services cover over 200 cities across China, surpassing 670,000 registered doctors and 45 million patients. Since its inception, the platform has facilitated over 700 million online health consultations for patients. Given its size, patient population, and geographic coverage, JianKang160 provides a comprehensive representation of the virtual health industry in the country.

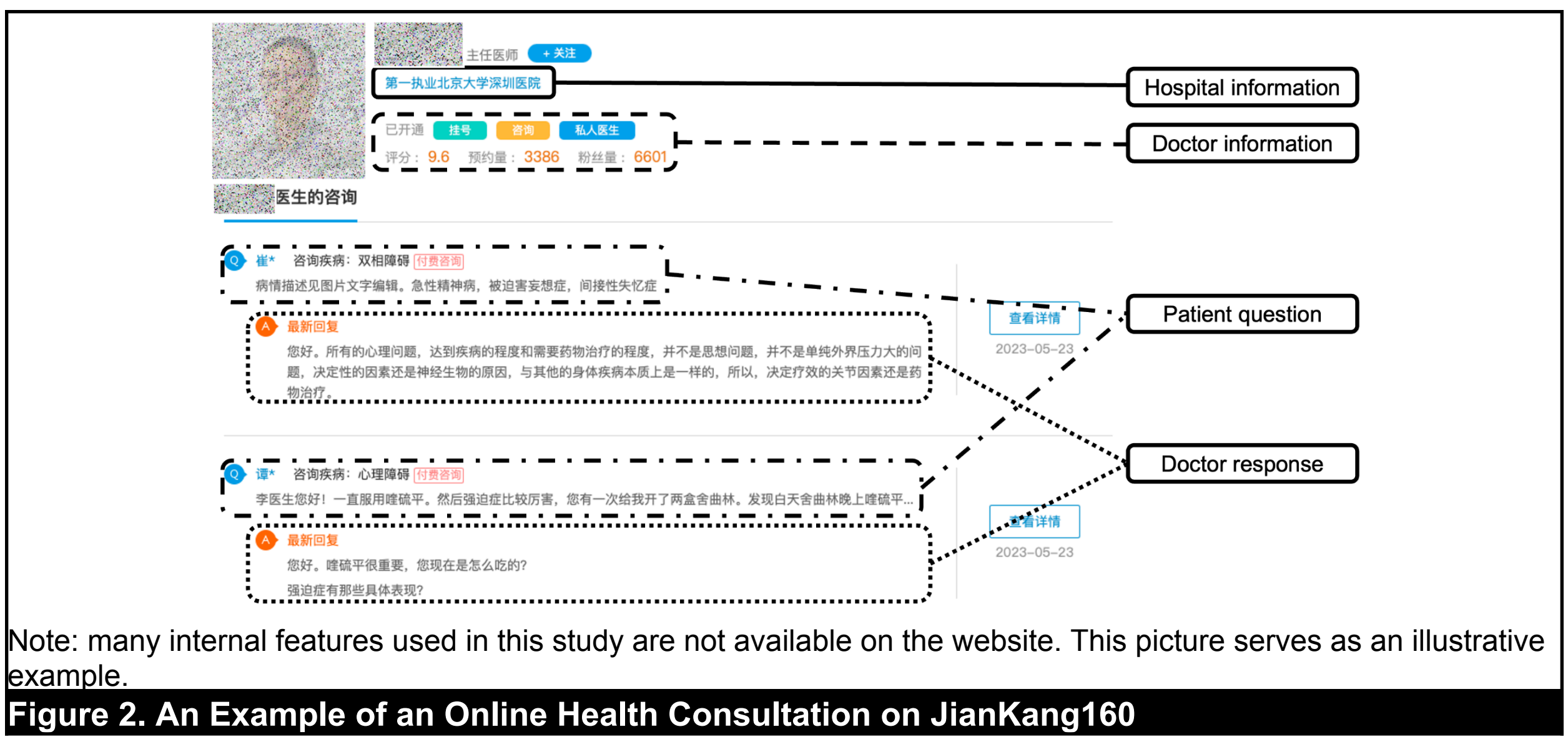

Note: many internal features used in this study are not available on the website. This picture serves as an illustrative example.

**Figure 2. An Example of an Online Health Consultation on JianKang160**

In this study, we focus on one of the major services of the platform: online health consultation, namely asynchronous doctor-to-patient consultation. Figure 2 shows an example of the doctor selection and consultation process. Table 1 shows the summary statistics of our research data. Through our collaboration and data confidentiality agreement with the platform, we have access to information about patients (de-identified), doctors, hospitals, offline hospital appointments, and online consultation dialogues. The patient and doctor information is recorded when they register for the platform. The online and offline appointment information is generated each time a patient requests an appointment, which records the patient's medical condition at each time. The dialogue information is the fine-grained descriptions of each online health consultation. The patients' offline appointment information is a unique asset of JianKang160, different from typical virtual health platforms that only have access to patients' online health consultation data. Such an advantage stems from JianKang 160's extensive offline hospital partner network, which is the largest in China among peers. These hospitals use JianKang160's cloud services to enable patients to make offline service appointments; therefore, online and offline data are shared mutually between the hospitals and the platform. These offline hospital appointment data can be linked to the patients, doctors, and online consultation data via unique patient and doctor IDs. The input data for our model includes structured, textual, and network data. In this study, we use data (with anonymized patient information) from all consultations that occurred between January 1, 2022, and June 20, 2022.

**Table 1. Summary Statistics of the Research Data**

| **Statistics** | | **Statistics** | |
|---|---|---|---|
| No. of consultations | 170,971 | No. of consultations per patient | 1.760 |
| No. of patients | 97,158 | No. of consultation per doctor | 24.414 |
| No. of doctors | 7,984 | No. of sentences per consultation | 10.491 |
| No. of hospitals | 956 | Average sentence lengths | 19.110 |
| No. of diseases | 80 | No. of offline hospital visits | 394,285 |
| **Doctor Titles** | **Percentage** | **Hospital Tiers** | **Percentage** |
| Chief physician | 24.94% | Specialized hospitals & centers | 48.2% |
| Associate chief physician | 30.97% | Third-level class A hospitals | 27.0% |
| Attending physician | 32.04% | Second-level class A hospitals | 8.3% |
| Physician | 9.74% | Second-level hospital | 6.4% |
| Psychotherapist | 1.20% | Third-level hospital | 4.8% |
| Traditional Chinese medicine | 0.70% | First-level hospital | 1.9% |
| Pharmacist | 0.41% | Other tiers | 3.4% |

## Unique Characteristics of the Data and Research Context

A prominent feature of our data is that it consists of both patients' online and offline healthcare digital traces. Patients' offline hospital appointment information goes beyond extant research on virtual health platforms that only have access to patients' online health consultation data. Such high-fidelity and wide range of internal data offer invaluable advantages for us to explore a more comprehensive picture of patients' entire healthcare journeys. Moreover, it affords us the opportunity to investigate whether this novel business model confers distinct advantages in the realm of virtual health, thereby warranting its promotion in more countries and regions.

In terms of the prediction task, while our goal is to predict the success of text-based online healthcare consultations, it is important to note that the input knowledge crucial for this task extends far beyond merely textual data. This complexity arises from the involvement of diverse stakeholders who are interconnected and engage in interactions that may influence one another during patients' healthcare journeys. Throughout patients' healthcare journey (both online and offline), interactions among stakeholders (e.g., patients, doctors, hospitals, and platforms) constitute a dynamic multi-view network (online view: online healthcare service, and offline view: in-person healthcare service): the nodes (stakeholders) whereby they may remain consistent, yet their online and offline links (interactions) vary. Furthermore, these stakeholders exhibit attributes that facilitate complex connections and information propagation (e.g., connections between patient characteristics, doctors' areas of specialization, and doctors' affiliated hospitals). These nuanced network features collectively inform the prediction of success in online healthcare consultations, presenting complexities that go well beyond textual data and the capacities of large language models (LLMs). We will further demonstrate in the evaluation section that LLMs (e.g., BERT and GPT-4o) cannot deal with our data types, as evidenced by their poor prediction performances.

Since the platform we examine is based in China, this introduces new characteristics to our research context. Unlike healthcare services in other countries (such as the US), healthcare services in China are highly transparent: (1) The pricing for individual medical services is transparent, with patients knowing the cost in advance. (2) The rank, experience, and professional level of doctors are transparent through their titles and the tiers of hospitals they are affiliated with. (3) Patients can transparently choose specialists, as there are normally no primary care physicians in China; most doctors have their own affiliated hospitals and are specialists, focusing on treating specific diseases. This system, different from the general practitioner model in many countries, provides us with the opportunity to explore the implications of such transparency and potential applications in other countries.

Another significant characteristic is that the task we aim to predict constitutes a new research problem. Our prediction focuses on the success of online text-based healthcare consultations on the virtual healthcare platform using comprehensive multimodal data generated during patients' healthcare journeys. This metric is a vital KPI closely monitored by our business partner as well as any other virtual health platforms in China (i.e., the labels are collected from patients and recorded in our business partner's internal health IT system). The online consultation service allows patients to engage in an unlimited number of online dialogues using text with the chosen doctor within 24 hours. The prediction task bears some resemblance to typical consumer transactions on e-retailers because, in this case, patients select a doctor, essentially a "product," based on various characteristics such as seniority levels, hospital affiliations, and specializations. However, there are differences as well. Patients are investing in a relationship and ongoing consultation process, expecting it to be successful. The uniqueness of this problem lies in the necessity to assess the likelihood of success at the outset of the consultation, presenting a distinct challenge from traditional transaction success prediction, where the prediction period is longer and allows for ample time to gather sufficient information. The focus here is to predict the outcome at the start of the consultation process, ensuring timely intervention, such as incentives to engage doctors and reminders if consultation success is not anticipated at the end of the 24-hour window. While it resembles a consumer product purchase with displayed prices and options, it is not a one-time purchase; rather, it marks the initiation of a 24-hour process. The application of this particular prediction problem can be extended to other virtual or hybrid business models, such as online education (e.g., predicting student retention at the beginning of a class), hybrid sales (e.g., predicting purchase intention while providing information online and offering offline testing and real-world experience), and more.

## HOMOPHILY THEORY AND KNOWLEDGE NETWORK CONSTRUCTION

Due to the digitalization of the healthcare industry, patients increasingly rely on digital technologies throughout the entire journey of seeking healthcare services. The digital traces of the healthcare journey are meticulously recorded by the healthcare IT infrastructure, which creates an intricate and dynamic network of connections involving patients (and their diseases), doctors (and their associated hospitals), and virtual health platforms. These stakeholders serve as entities within the network. The relationships among the stakeholders are through the patients' healthcare journey, the doctors' affiliation with the hospitals, as well as doctors' expertise in specific diseases. Moreover, these relationships further propagate to other entities in the network, such as patient and patient (e.g., patients with the same disease forming the disease co-occurrence network), doctor and disease (e.g., doctors specializing in treating specific diseases), doctor and hospital (e.g., doctors being affiliated with hospitals).

Homophily theory, a cornerstone in sociology, posits that individuals with similar characteristics (e.g., demographics and medical history of patients) or interests (e.g., disease treatment) in a relationship network tend to gravitate toward each other and behave in similar ways (Ertug et al., 2022). Homophily-related studies can be categorized into two broad categories: individual (i.e., choice homophily) and structural (i.e., homophily that is induced by the structures of opportunity and constraint). This work employs structural homophily, which concerns the alignment of opportunities (e.g., the ability to access optimal or the most suitable services required by the patient on the platform) and constraints (e.g., a patient having a specific medical condition and thus limited to seeking assistance from a particular subset of doctors) within a relational network formed by the digital traces of patients' healthcare journeys. In this network of relationships, the similarity among patient nodes (e.g., node positions, connections with other entities, propagation of patient nodes to other entities such as disease-to-disease networks, network topology, and dynamics) reflect and assist in predicting patient behaviors and actions (e.g., success in online consultations).

**Table 2. Theory-driven Knowledge Network Construction**

| Motivation | | Reference | Research Gap | Our Design |
|---|---|---|---|---|
| Homophily theory | The homophily of a network can be leveraged to enhance network-related predictive analysis. | (H. Chen & Deng, 2023) | In the context of virtual health predictive analysis, patient information proves to be extremely sparse and incomplete. | Implicit knowledge extraction from networks formed by patient digital traces |
| | In a relationship network, the attributes of entities influence the formation of homophily within the network. | (McPherson et al., 2001) | The current knowledge network does not take into account attribute encoding and the interaction effects of attributes. | Attribute encoding of knowledge network |
| | In a relationship network, the evolving nature of the network (e.g., the joining of entities, the formation of links between them) contributes to the development of homophily within the network. | (Kossinets & Watts, 2009) | The current knowledge network does not account for network evolution. | Dynamic knowledge network |
| With the digital transformation taking place across various industries, the hybrid mode has emerged as a new phenomenon and trend. | | (G. Chen et al., 2023) | The current knowledge network does not take into consideration the multi-view nature (online and offline) of the network. | Multi-view knowledge network |

Driven by the homophily theory, within the proposed framework, we construct a knowledge network to capture the *implicit knowledge* (i.e., the network formed with digital traces in patients' healthcare journeys) to enhance the downstream online consultation success prediction performance (Table 2). The network has the following three characteristics:

(1) The network has multiple views. Patients can seek medical services through online platforms or in-person healthcare facilities; doctors can offer their services on online healthcare platforms or be affiliated with in-person healthcare facilities. The multiple views (online and offline) are intertwined and mutually influential, holding significant implications for understanding patients' healthcare journeys. In addition, online healthcare consultation and offline health services are complementary to each other: virtual health services have unique advantages when additional test results are not needed, immediate medical advice is required, the patient needs further clarification from the doctor regarding the need to visit the hospital, or a second opinion is necessary for a known diagnosis or test results. This indicates that patients' offline hospital appointments and online consultation records are interconnected. Integrating patient online and offline medical records can further alleviate the data incompleteness in previous studies that consider patient online experiences alone. Therefore, we construct a knowledge network, $G$, with two cohesively correlated views, denoted by $G_{online}$ and $G_{offline}$, according to online and offline digital traces of patient healthcare journeys. These two views of $G$ share information for the same set of entities (i.e., patient, doctor, disease, and hospital entities). Nevertheless, their connectivities and topological structures are distinct because they are uniquely derived from online and offline information.

(2) The entities, their attributes, relationships, and the propagations of the entities' relationships contain latent information that can assist in predicting healthcare consultation success during online healthcare services. Figure 3 provides two examples. In Example 1: the indirect connections between entities enable the exploitation of higher-order implicit relationships between two unconnected entities, such as between "Hospital 1" and "Hospital 2" (Figure 3). Fully exploiting the higher-order relations within the network is crucial for understanding patients' expectations and needs, which can contribute to healthcare consultation success prediction. In Example 2: connectivities between disease nodes in the network are effective in capturing diseases' co-occurrence relationships and can enrich extant patient-disease and doctor-disease relationships (Figure 3), which are critical for evaluating the patient-doctor interaction quality and predicting healthcare consultation success.

(3) The network is dynamic and evolves over time, reflecting changes in the healthcare journeys of various patients. Since our goal is to predict the healthcare consultation success of an online consultation, certain entities and specific relations among them play a crucial role in the particular prediction and require the utmost attention during the prediction process. Moreover, it is also important to consider the other connections and historical relationships of these entities within the network, as they can provide latent information for healthcare consultation success prediction. The incorporation of an attention mechanism is imperative to discern the entities and relationships existing within this dynamic and intricate network, pinpointing those that demand heightened focus during the ongoing prediction phase.

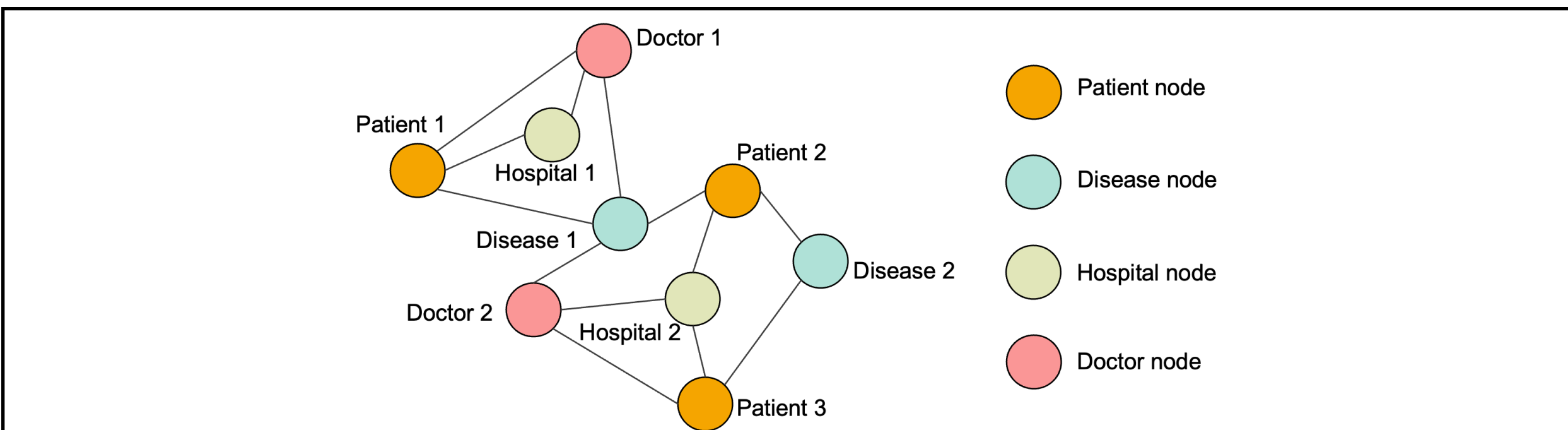

**Example 1**:
Path: *Hospital 1 – Patient 1 – Disease 1 – Patient 2 – Hospital 2*
Explanation: *Patient 1* is treated by a doctor affiliated with *Hospital 1*; *Patient 1* has *Disease 1*; *Patient 2* has *Disease 1*; *Patient 2* is treated by a doctor affiliated with *Hospital 2*.
Implication: The tiers of hospitals may reflect the patient's expectations regarding the quality of the doctor's services. The relationships or tiers of Hospital 1 and Hospital 2, along with their respective hospital-level consultation success rates, are important factors to consider when predicting the likelihood of success during the current consultation.

**Example 2**:
Path: *Patient 1 – Disease 1 – Patient 2*
Explanation: *Patient 1* has *Disease 1*; *Patient 2* has *Disease 1*.
Implication: The ease of diagnosing and treating Disease 1, along with the overall consultation success rate of Disease 1, are important factors to consider when predicting the likelihood of success during the current consultation.

**Figure 3. An Example of the Knowledge Network**

Based on the structural homophily theory, we consider that the structures of opportunity and constraint of four types of entities in patients' healthcare journeys contain important latent information – patients, doctors, hospitals, and diseases – which can benefit the downstream patient consultation success prediction. Patients and doctors are identified as network entities as they represent key stakeholders in online healthcare consultations. Hospitals are identified as entities because, in our context, hospitals can list their own information on the platform, which allows patients to make appointments for in-person visits. Alternatively, patients can request an instant online consultation service from doctors at the participating hospitals. Therefore, hospitals serve as an indispensable link between patients and doctors. We also include disease as another entity type because, besides the interactions between the three key stakeholders (hospital, patient, doctor), disease entities and their relationships with stakeholders contain important medical knowledge such as disease cooccurrences, doctors' and hospitals' specialized diseases, and patient healthcare preferences. Overall, we account for four entities (i.e. patient-, doctor-, hospital-, and disease-entities) and five types of relationships in the knowledge network as listed in Table 3.

**Table 3. Entities and Relationships Defined in the Medical Knowledge Network**

| | |
|---|---|
| Entities | Patient ($e_{patient}$), doctor ($e_{doctor}$), hospital ($e_{hospital}$), disease ($e_{disease}$). |
| Relationships | Doctor-Hospital relationship ($r_{doc_hosp}$), Doctor-Disease relationship ($r_{doc_dis}$), Patient-Disease relationship ($r_{pat_dis}$), Patient-Hospital relationship ($r_{pat_hosp}$), Patient-Doctor relationship ($r_{pat_doc}$). |

Note: In our research setting, we define the knowledge network as an undirected network, where the relationship denotes the existence of a connection between two entities. We disregard the direction of the connection between entities because both the flow of information from and to entities are crucial for constructing the knowledge network. For instance, "doctor→hospital" indicates that a doctor is affiliated with a hospital, and the tier of the hospital may imply the patient's expectations regarding the doctor's services. Conversely, "hospital←doctor" represents a relationship between a hospital and a group of doctors. Sometimes, the overall online service level of doctors in a hospital may be higher than that of doctors in other hospitals of the same tier due to training, which also impacts patient online consultation success prediction.
To summarize, in our research setting, the two-way dissemination of information holds equal importance in knowledge network construction and patient consultation success prediction. Therefore, our knowledge network is an undirected network.

## RELATED WORK

### Language Model for Explicit Knowledge Representation

One distinctive feature exclusive to online healthcare consultations is the meticulous documentation of the patient-doctor communication process, including every choice of words, sentence construction, intervals between dialogues, and more. This information constitutes *explicit knowledge* that directly reflects the various needs of patients at different levels and the corresponding responses from doctors, including but not limited to disease inquiry needs, knowledge acquisition needs, and emotional support needs. Such explicit knowledge plays a crucial role in determining the success of a consultation (the formal definition is in Section: Problem Formulation), thereby influencing healthcare consultation success. *Our first research question is: how can we effectively represent explicit knowledge to improve the prediction success of online health care consultation?*

In the past decade, pre-trained language models have propelled natural language processing (NLP) into a new era (Qiu et al., 2020). Well-trained language models are designed to capture semantic information (Vulić et al., 2020); in our context, they need to comprehend the patient's diagnosis inquiries and knowledge acquisition during patient-doctor communication, as well as the extent to which doctors' responses align with these needs, to identify patterns (e.g., the matching level of language vector representations between patients and doctors). Meanwhile, well-trained language models excel at distinguishing similar content with different emotional expressions (Zhou et al., 2023); in our context, they are expected to be adept at capturing the characteristics of common emotions and sentiments that indicate successful communication between doctors and patients and comparing these expressions with characteristics in failed communication processes, thereby revealing patterns.

In the diverse landscape of existing language models, we aim to adapt a model that has been well-established in previous research for its high performance. Current state-of-the-art language models can be categorized as encoder models (e.g., the BERT family), decoder models (e.g., the GPT family), and sequence-to-sequence models (e.g., the BART family) (Wolf et al., 2020). Encoder models are best suited for tasks requiring an understanding of the full sentence, such as document classification. Decoder models usually revolve around predicting the next word in a sentence, making them best suited for tasks involving text generation. Sequence-to-sequence models are ideal for tasks involving the generation of new sentences based on a given input, such as document summarization (Wolf et al., 2020). Given that our prediction goal (i.e., online healthcare consultation success) is a binary classification problem, the BERT family is best suited for our specific prediction task. Moreover, the chosen model should exhibit proficiency in processing Chinese (i.e., our business partner is located in China and Chinese is the primary language of communication on the online healthcare consultation platform). We are located in the MacBERT (i.e., the pre-trained model for Chinese natural language processing) (Cui et al., 2020). Specifically, we aim to achieve better performance in healthcare consultation success prediction tasks by effectively representing *explicit knowledge* (i.e., patient-doctor consultation dialogues) through fine-tuning the MacBERT model.

The current MacBERT model can fulfill most of our requirements to represent *explicit knowledge* of online healthcare consultation. However, there is another aspect that holds significance in our context and prediction tasks: the time intervals between dialogues. These intervals are crucial as they directly reflect the timeliness of a doctor's response and the smoothness of communication between the doctor and the patient. Therefore, we aim to devise a new language model representation that differs from the existing MacBERT, as it can capture the time intervals between dialogues.

### Knowledge Graph Attention Networks for Implicit Knowledge Mining

While *explicit knowledge* plays a crucial role in predicting the success of an online healthcare consultation, the multi-view dynamic network formed by the interactions among various stakeholders throughout patients' healthcare journey can also significantly contribute to forecasting consultation success and addressing the challenges of data sparsity and incompleteness inherent in virtual contexts.

As mentioned earlier, guided by homophily theory, we aim to capture valuable *implicit knowledge* in the dynamic multi-view network including the attributes of various entities, their higher-order propagations, and the intertwined influences among them. Directly obtaining and incorporating such *implicit knowledge* into the prediction model is challenging. For example, directly incorporating stakeholders' attribute information into the predictive model cannot capture high-order propagated information. On the other hand, there are multiple stakeholders in the network, and each stakeholder has an impact on healthcare consultation success prediction; learning the high-order representation of individual stakeholders would necessitate multiple models with multiple training, which is inefficient and would result in the loss of their interrelated relations. Overall, comprehending and extracting *implicit knowledge* from this dynamic multi-view network is meaningful for predicting healthcare consultation success in the virtual context. *Our second research question is: how can we efficiently represent implicit knowledge in the multi-view dynamic networks to improve the prediction success of online health care consultation?*

Knowledge networks, an increasingly popular research direction, are interconnected systems that organize and link entities, representing structural relations between entities (Ji et al., 2022). Knowledge networks allow for efficient retrieval and mining of knowledge, as well as enhancing decision-making processes (Phelps et al., 2012). Broadly speaking, research on knowledge networks can be classified into four main groups: knowledge acquisition, knowledge representation learning, temporal knowledge networks, and knowledge-aware applications (Ji et al., 2022). In this study, we aim to utilize a knowledge network to learn representations of stakeholders and accomplish three objectives: 1) represent various stakeholders within multiple views (i.e., online healthcare and in-person healthcare); 2) extract the complex and dynamic relationships among the stakeholders; 3) utilize these stakeholder representations to enhance the prediction of healthcare consultation success in the virtual context. As a result, our study lies at the intersection of knowledge representation learning, temporal knowledge networks, and knowledge-aware applications among the four classes of knowledge networks.

However, the current knowledge network representation is insufficient to meet our objectives. The first challenge is that there are multiple stakeholders (i.e., patients and their associated diseases, doctors and their associated hospitals, and platforms), and they are associated with rich attribute information relevant to their representation learning. However, extant knowledge networks fall short of incorporating entity attribute information in representation learning. Even though a few studies linearly combine initial features after principal component analysis and attach feature embeddings to the entity (Zheng et al., 2021), they are limited by linearity and cannot capture the non-linear inherent feature interactions and the inherent structure. Effectively integrating attribute information into the knowledge network while enabling attribute information propagation between neighboring entities is non-trivial.

The second challenge stems from the evolutionary nature of patients' healthcare journeys and stakeholder relationships. As patients' health condition progresses, patients' consultation needs change accordingly. At different time points, the interactions between stakeholders (i.e. Patient-Doctor relationship ($r_{pat_doc}$), Patient-Disease relationship ($r_{pat_dis}$)) represent a distinct snapshot of the patient's health condition at that stage. These interactions are of different importance for the extraction of patients' needs at different time periods. Although the time-aware attention mechanisms in existing work can account for the uneven contribution of each interaction in a behavior sequence, they focus more on the immediate and salient features and fail to learn the long-term dependencies in the sequence and the complex stakeholder relationships underlying these interactions. Extant static knowledge networks can capture the dependencies between different interactions and implicit high-order relationships between stakeholders. However, for predictions at different time points, attention and weights should be estimated dynamically for a specific time segment of the knowledge network. Therefore, capturing the dynamic importance of different sub-networks for prediction necessitates careful model design.

To overcome the two technical challenges, we build upon prior research in attention-based knowledge networks (Wang et al., 2019), with the objective of devising a novel knowledge network that distinguishes itself from existing work by its capacity to represent: a) multi-view perspectives; b) the encoding and propagation of attributes associated with nodes (e.g., stakeholders); c) network dynamics contributing to the ultimate healthcare consultation success in the context of online healthcare consultations.

## Multimodal Fusion of Explicit and Implicit Knowledge

We have two critical types of knowledge that are important to healthcare consultation success in the virtual context. (1) *Explicit knowledge*: patient-doctor communication process in text format.

They uncover how patients have interacted with healthcare providers and their level of satisfaction with the care they received, which are critical for predicting healthcare consultation success. (2) *Implicit knowledge*: the multi-view dynamic network incorporates information from patients, doctors, hospitals, and virtual health platforms, obtained through the digital traces of patient healthcare journey. Such a network provides essential implicit knowledge, including the doctor's expertise, service characteristics, patient health status, previous treatments, medical conditions, preferences, and so on. This information is also valuable for accurately predicting healthcare consultation success. Given that *explicit knowledge* and *implicit knowledge* inherently belong to disparate feature spaces and convey distinct predictive features of healthcare consultation success, a simple feature combination is not feasible. *Our third research question is: how can we seamlessly integrate implicit and explicit knowledge to improve the prediction success of online health care consultation?*

In many problem domains, information regarding the same prediction objective can be obtained from different sources or under various conditions (i.e., modalities). In these problem domains, given the rich characteristics and complexity of the prediction goal, it is uncommon for a single information modality to provide complete knowledge of the goal. When multiple modalities are available for the same problem, they introduce new dimensions of information and complexity, raising possibilities and challenges beyond those related to the individual exploitation of each modality (Lahat et al., 2015). Accordingly, multimodal data learning has emerged as a rapidly expanding research field (J. Gao et al., 2020). In general, multimodal data learning can be categorized into the following categories (please note that there may be overlap in different areas, and some studies may belong to multiple categories). (1) Deep learning architecture engineering for multimodal learning: designing and optimizing neural network models that can effectively process and integrate information from multiple data modalities. For example, Lu et al. (2019) propose ViLBERT to learn task-agnostic joint representations of image and text content by extending the BERT architecture to a multi-modal two-stream model. (2) Multimodal fusion: the process of integrating information from multiple types of sensory modalities to improve the performance and robustness of machine learning models. For example, Y. Gao et al. (2016) propose compact bilinear representations with strong discriminative power and relatively low dimensions. (3) Cross-modality learning: leveraging information from one modality to improve the understanding and performance in another modality. For instance, Li et al. (2019) propose the VisualBERT framework for modeling a broad range of vision-and-language tasks. (4) Shared representation learning: the process of learning a common or joint representation for data from multiple modalities. For example, Radford et al. (2021) propose the CLIP (contrastive language-image pre-training) model to align textual descriptions with images in a shared representation space using a contrastive learning approach.

In this work, we aim to fuse *explicit knowledge* (i.e., patient-doctor communication processes) and *implicit knowledge* (i.e., the multi-view dynamic network formed through digital traces of patients' interactions with various stakeholders during their healthcare journey) to predict the success of online healthcare consultations. Our proposed fusion method falls under the category of multimodal fusion for the following reasons. (1) As mentioned in the previous sections, there exist powerful and well-suited methods for generating representations of *explicit knowledge* (i.e., language models) and *implicit knowledge* (i.e., knowledge networks). Therefore, designing a different deep learning architecture solely for multimodal learning purposes is not the most efficient way and may not be able to extract the *explicit and implicit knowledge*; hence, we do not choose the route of "Deep learning architecture engineering for multimodal learning." (2) Although both *explicit and implicit knowledge* significantly contribute to predicting the success of online healthcare consultations, the information implied between these two modalities cannot interact during the representation generation stage. This is because *explicit knowledge* is generated from language models, while *implicit knowledge* is extracted from knowledge networks. In addition, their source data represent different types of information. Cross-modality learning and shared representation learning methods tackle a different scenario – they typically process image and text modalities, where both convey the same information. In such a case, text and image can be fused early during the representation generation phase. Considering the above-mentioned four categories of multimodal data learning approaches as well as these two restrictions, we resort to the multimodal fusion technique to fuse representations generated from language models and knowledge networks.

In our research context, there is still a challenge revolving around the efficient fusion of multimodal information derived from language models and knowledge networks. On one hand, retaining crucial information from diverse modalities poses a challenge when directly concatenating all vector representations and their interactions (e.g., representation multiplication), as it results in a substantial escalation of representation dimensions. For instance, in our context, the concatenation and multiplication of representations of all modalities amounts to 1,056 and 278,784 dimensions, respectively, which poses severe challenges for downstream predictions. On the other hand, reserving useful multimodal information while mapping vector representation interactions to low-dimensional space using a mapping algorithm would inevitably cause a loss of information. In addition, representation distributions may vary among different modalities. Consequently, a single mapping algorithm may only suit a single modality and fail to capture sufficient information from other modalities.

This study aims to expand upon prior research on multimodal data fusion (Chetouani et al., 2020; López-Sánchez et al., 2020; S. Xu et al., 2023) by developing a novel fusion mechanism that distinguishes itself from existing approaches in two key aspects: (1) taking into account the unique data distributions inherent in various data modalities and (2) striving to achieve a balance between the efficiency of data fusion and the creation of a comprehensive representation.

## Key Novelties of Our Study

Our proposed framework offers several key contributions. First, we introduce an innovative knowledge network that stands out from existing models through its ability to represent: multi-view perspectives of the knowledge network, the encoding and propagation of node attributes, and network dynamics. Second, we present a new multimodal data fusion method that is unique in accounting for the distinct data distributions present in various data modalities and achieving a balance between efficient data fusion and thorough representation. The collective contributions outlined herein significantly augment predictive capabilities in the context of online healthcare consultations. Furthermore, these contributions advance the methodology within the field of design science, offering potential applications in diverse problem domains. This extends to scenarios characterized by the coexistence of explicit and implicit knowledge, where the amalgamation of these two forms of knowledge holds the promise of enhancing predictive analysis.

## RESEARCH DESIGN

### Problem Formulation

We observe a patient base $P$, encompassing their online and offline health digital traces on a healthcare service platform that provides online health consultation services and offline doctor appointment services. For each patient $p \in P$, we collect his or her online consultation traces $c_{online} = \left(c^{1}_{online}, c^{2}_{online}, ..., c^{N_p}_{online}\right)$ and offline consultation traces $c_{offline} = \left(c^{1}_{offline}, c^{2}_{offline}, ..., c^{N_p}_{offline}\right)$, where $N_p = \{1, 2, ..., i, ...\}$, ordered in time. Each online consultation record $c^{i}_{online}$ or offline consultation record $c^{i}_{offline}$ corresponds to a patient ($e_{patient}$) - hospital ($e_{hospital}$) - doctor ($e_{doctor}$) - disease ($e_{disease}$) interaction relationship, denoted by $r^{i}_{online}$ or $r^{i}_{offline}$, where $r^{i}$ consists of Doctor-Hospital relationship ($r_{doc_hosp}$), Doctor-Disease relationship ($r_{doc_dis}$), Patient-Disease relationship ($r_{pat_dis}$), Patient-Hospital relationship ($r_{pat_hosp}$), and Patient-Doctor relationship ($r_{pat_doc}$). For an online consultation $c^{i}_{online}$, we also observe the patient-doctor consultation dialogues comprising a sequence of sentences $l^{i}_{online} = \left(l_1, l_2, ..., l_{m_p}\right)$. Given a patient's historical interaction relationships $r_{online} = \{r^{1}_{online}, ..., r^{i}_{online}, ..., r^{N_p}_{online}\}$ and $r_{offline} = \{r^{1}_{offline}, ..., r^{i}_{offline}, ..., r^{N_p}_{offline}\}$ and the consultation dialogues of online consultation $l^{i}_{online}$, we aim to design a new predictive model $f_{\Phi}$ that captures essential information from those relationships (*implicit knowledge*) and consultation dialogues (*explicit knowledge*) and predicts whether that patient $p$' online consultation is successful. The ground truth label (success vs failure) is collected by our business partner and recorded in their IT database. Let the prediction outcome be $\tilde{y}_c = \{0, 1\}$, where 1 indicates that patient $p$'s consultation is failed. Formally, the healthcare consultation success prediction problem is a binary classification problem, formulated as:

$$\tilde{y}_c = argmax\, p\left(y_c | f_{\Phi}\left(r_{online}, r_{offline}, l^{i}_{online}\right)\right) \quad (1)$$

## Consultation Success Prediction Framework

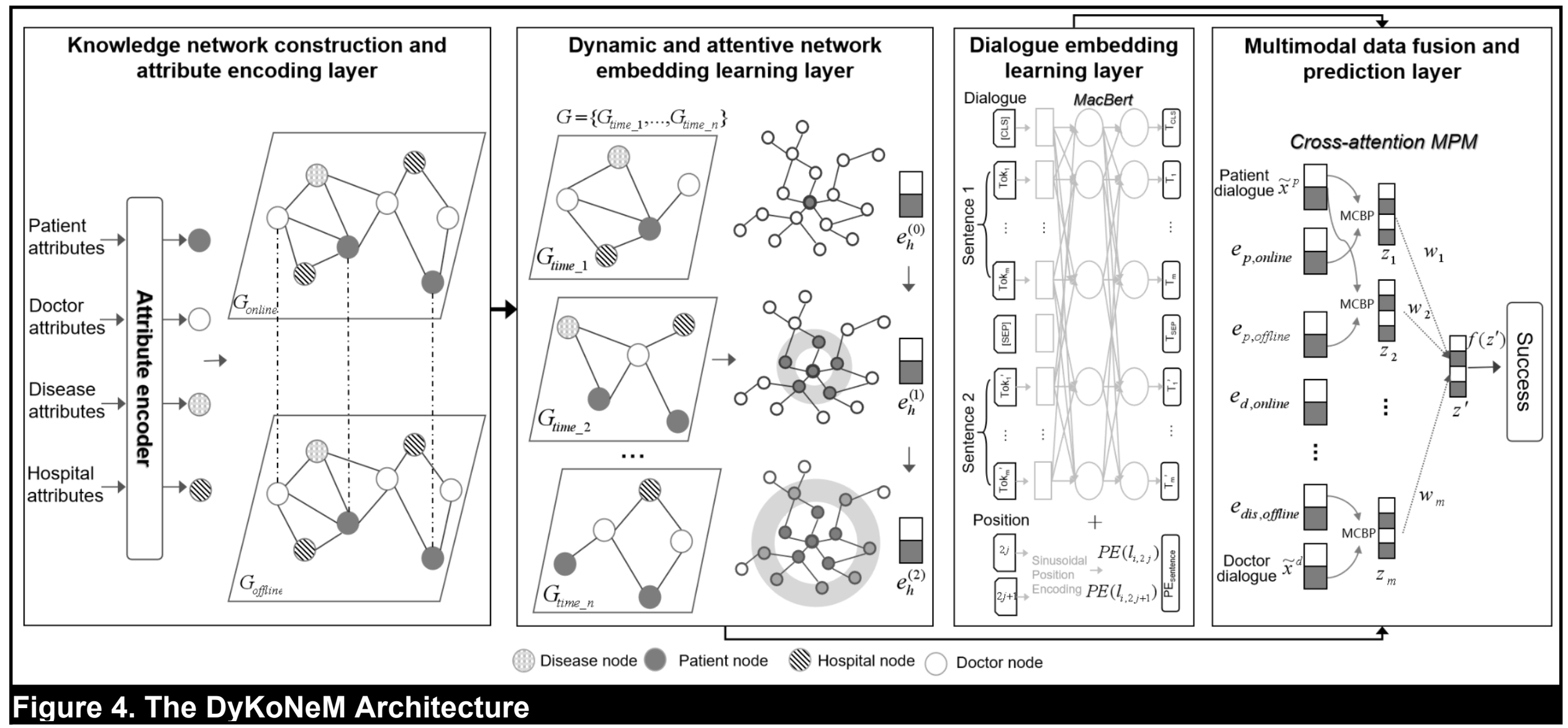

**Figure 4. The DyKoNeM Architecture**

We develop a Dynamic Knowledge Network and Multimodal Data Fusion (DyKoNeM) consultation success prediction framework (Figure 4). Our main motivation is to develop a novel deep mining model coupled with implicit (network formed with digital traces in patient healthcare journey) and explicit knowledge (patient-doctor communication process) that are essential for accurate healthcare consultation success prediction in the virtual context.

DyKoNeM features four layers: deep bidirectional transformer for explicit knowledge extraction, knowledge network construction with attribute encoder for multi-stakeholder feature extraction, dynamic and attentive temporal information propagation and aggregation for stakeholder representation learning, and multimodal data fusion for feature fusion and success prediction. The first layer encodes the patient-doctor dialogue information from the online consultation. The second layer extracts implicit relationships among stakeholders and their features in the virtual health system. The third layer derives the dynamic and uneven contributions of stakeholder relationships to healthcare consultation success prediction. The fourth layer takes into account the implicit stakeholder relationship knowledge extracted from both patient online consultations and offline in-person hospital visits, and the explicit patient-doctor communication dialogue knowledge with a fusion mechanism for healthcare consultation success prediction.

**Dialogue Embedding Learning Layer:** The design of this layer aims to capture explicit knowledge about the patient-doctor communication process and interaction quality. They contain critical information about patient health needs and doctor responses, which are critical indicators of predicting consultation success. As mentioned in the related work, we locate MacBERT as the embedding approach for the dialogues, which is fine-tuned for Chinese NLP tasks (Cui et al., 2020). Unlike the existing MacBERT, we also incorporate an additional position encoding method (Vaswani et al., 2017) to extract the dialogue sequential features (i.e., the time interval between dialogues) that reflect the interplay between patients and doctors as they directly reflect the timeliness of a doctor's response and the smoothness of communication between the doctor and the patient. Let $\left(l_1, l_2, \ldots, l_m\right)$ denote the sequence of patient or doctor dialogue sentences. The dialogue embeddings for patients and doctors are obtained using the same approach:

$$\tilde{x}_i = MacBERT\left(l_i\right) \tag{2}$$

$$PE(l_{i,2j}) = sin\left(\frac{i}{10000^{2j/d}}\right) \tag{3}$$

$$PE(l_{i,2j+1}) = cos\left(\frac{i}{10000^{2j/d}}\right) \tag{4}$$

$$\tilde{x} = \frac{1}{m}\sum_{1}^{m}\left(\tilde{x}_i + PE(l_i)\right) \tag{5}$$

where $i$ is the sentence position in the dialogue, $j$ is the dimension, $PE(l_{i,2j})$ and $PE(l_{i,2j+1})$ are the sinusoid encodings for the $2j^{th}$ and $(2j + 1)^{th}$ dimension of the position encoding.

**Knowledge Network Construction and Attribute Encoding Layer:** The design of this layer aims to model the networks that patients formed through digital traces of interactions with various stakeholders during their healthcare journey. The knowledge network $G$ and the two views of this network, $G_{online}$ and $G_{offline}$, can be formally defined as $\{(h, r, t)|h, t\epsilon E, r\epsilon R\}$, where each triplet indicates that there is a relationship $r$ from head entity $h$ to tail entity $t$. Nevertheless, the connectivities and topological structures are distinct for these two views of the knowledge network, as they are uniquely derived from online and offline patient healthcare journeys. Notably, the interactions among the entities in the networks are dynamic in nature, which motivates us to assign a timestamp $t$ for each knowledge network according to its period of observation. An example of $G_{online}$ is depicted in Figure 3.

Given a knowledge network, $G$, with a set of doctor-, patient-, hospital-, and disease-entities and entity relations, the attribute encoding layer encodes heterogeneous entity attributes, such as doctor title, hospital tier, patient age, among others, to enrich the node embeddings and enable indirect interactions among node attributes. Compared to the conventional random homogeneous initial embedding approach (Breit et al., 2020; Mohamed et al., 2020) which leads to slow model convergence and sub-optimal node representations, and linear feature encoding approaches (Zheng et al., 2021) which fall short in exploiting node attribute interactions, we propose an attribute interaction-based encoding mechanism for initialization to effectively inject attribute information into the heterogeneous knowledge networks.

Specifically, let $e_{id}$ be the node embedding vector for entities in the heterogeneous knowledge networks, $A$ be the set of node attributes, and $E_{att} \epsilon R^{k\times e}$ be the attribute embedding matrix for the lookup of a node's attribute multi-hot representations. The attribute encoding layer first computes the interactions between the node's attributes and linearly combines the multiple interactions. Next, a fully connected layer is deployed to obtain the high-order attribute representation:

$$e_{att} = LeakyReLU\left(w_1 \Sigma_{ai,aj\in A}\left(e_{ai} \odot e_{aj}\right)\right) + b_1 \quad (6)$$

where $e_{ai}$ and $e_{aj}$ are attribute multi-hot representations retrieved from $E_{att}$, $w_1$ is a trainable parameter matrix, and $b_1 \in R^e$ is the bias. This high-order attribute representation contains not only individual attribute information but also the co-occurrence relationships between multiple attributes. We then concatenate the high-order attribute representation with the initial node embedding, and feed it into a fully connected layer to get the encoded node embedding with attribute information:

$$\widetilde{e} = LeakyReLU\left(w_{p2}\left[e_{att};\ e_{id}\right]^{\top}\right) + b_{p2} \quad (7)$$

where $w_2$ is a trainable parameter matrix, and $b_2 \in R^e$ is the bias. For entities with few entity neighbors and relations in the heterogeneous knowledge network, the attribute encoder helps generate informative embeddings from their attributes, enables the propagation of attribute information along the network path quality, and enhances the quality of representation learning. The $G_{online}$ and $G_{offline}$ entity embeddings will be utilized as the feature fusion module for healthcare consultation success prediction. The attribute interaction encoding is flexible to integrate extra and heterogeneous node attributes, enabling future knowledge network extension.

Compared to existing attribute integration approaches in knowledge networks (Breit et al., 2020; Mohamed et al., 2020), the main difference and benefit of our attribute encoder lie in its capability to learn implicit and non-linear interactions between node attributes and generate more meaningful initial embeddings for knowledge network training. Notably, initial embeddings with richer information can enhance the embedding quality of rare entities that cannot receive sufficient updates in the learning process, and provide valuable insight into the formation of homophily within the network.

**Dynamic and Attentive Network Embedding Learning Layer:** The design of this layer aims to explicitly model the dynamic nature of the above knowledge network. This is because patient-doctor-disease-hospital relations evolve over time. Feeding the entire set of online health consultation records into a single static knowledge network neglects the dynamic nature of knowledge and may fail to capture the complex relationships of different stakeholders for online health consultation success prediction. To tackle this issue, we first design a sliding time window-based temporal knowledge aggregation mechanism, which divides the entire observation period into $m$ equal-length time windows, and subsequently obtains a set of sub-knowledge networks $\left\{G_1, G_2,..., G_m\right\}$. Local entity embeddings in sub-networks are learned only from observations happening in the corresponding time window. For example, if a patient only has medical records in the $k^{th}$ and $m^{th}$ time period, his or her local embeddings are learned from $G_k$ and $G_m$. Meanwhile, the patient's global entity embedding is learned from the global knowledge network $G$. The local entity embeddings learned from different time periods may have uneven contributions to healthcare consultation success prediction. More recent information is usually of greater importance for implicit knowledge learning. Therefore, we introduce a learnable decay factor $\beta \in R$ to discriminate the importance of different subnetworks in $\{G_1, G_2, ..., G_m\}$. The importance of a subnetwork $G_k$ is defined as:

$$w_{importance} = \frac{e^{-\beta \cdot (m-k)}}{\sum_{k=1}^{m} \left(e^{-\beta \cdot (m-k)}\right)} \tag{8}$$

where a larger $k$ corresponds to a more recent time period. Next, we fuse the local entity embeddings with global entity embedding:

$$e^{local} = \sum_{j=1}^{m} w_{importance} e^{j} \tag{9}$$

$$e^{overall} = LeakyReLU\left(W_{temp}\left[e^{local}, e^{global}\right]^{\top} + b_{temp}\right) \tag{10}$$

where $e^{local}$ denotes the unified local embeddings, $e^{global}$ is the global embedding, $e^{overall}$ represents the entity embedding containing both local and global implicit knowledge for online consultation success prediction, $W_{temp}$ is a trainable parameter matrix, and $b_{temp}$ is the bias parameter.

The learning of entity global embedding $e^{global}$ and local embedding $e^{local}$ comprises the same processes: information propagation and aggregation, and embedding decoding. The information propagation and aggregation along the relations in the heterogeneous knowledge network is key to the exploration of high-order implicit relationships between entities. Entities can recursively absorb information from neighbors through multi-layer information propagation. As noted earlier, essential attribute information can propagate along the pathway in the knowledge network, which is important for fully exploiting the complex knowledge network and acquiring comprehensive entity representations. In our heterogeneous knowledge network, different entity neighbors make uneven contributions to the target node embeddings. For example, a patient can have multiple diseases, doctors, and hospital entity neighbors in the network. Disease and doctor entity neighbors should have more influence on the patient's embedding than hospital entity neighbors because a hospital entity contains diverse medical relations and has more general embeddings. Therefore, we discriminate the information contribution among entity neighbors by employing an attention weight in the information propagation and aggregation (Wang et al., 2019). To be specific, given a set of triplets $(h, r, t) \epsilon N_h$, where $h$ is the head entity, attribute node embedding $e_h$ and $e_t$ are the head entity and tail entity produced by the attribute encoder, entity neighbors with first-layer connections with the head entity aggregate their information through a weighted linear combination of tail embeddings:

$$e_{N_h} = \sum_{(h,r,t) \epsilon N_h} \pi(h, r, t) e_t \tag{11}$$

where $\pi(h, r, t)$ represents the knowledge-aware attention weights allocated to each entity neighbor, which controls the amount of information propagated from the tail entity $t$ via relation $r$. $\pi(h, r, t)$ depends on the distance between $e_h$ and $e_t$ in $r$'s relation space and is formulated as:

$$\pi(h, r, t) = \left(W_r e_r\right)^{\top} tanh\left(W_r e_r + e_r\right) \tag{12}$$

$$\pi(h, r, t) = \frac{exp(\pi(h,r,t))}{\sum_{(h,r',t') \epsilon N_h} exp(\pi(h,r',t'))} \tag{13}$$

Next, $e_{N_h}$ is fused with the current node embedding $e_h$ to obtain the first-layer information aggregated embedding $e_h^{(1)}$ for entity $h$. The fusion process employs the bi-interaction aggregation operation to consider two types of feature interactions between $e_h$ and $e_{N_h}$, and feed the output into a fully connected layer:

$$e_h^{(1)} = f\left(e_h, e_{N_h}\right) = LeakyReLU\left(W_1\left(e_h + e_{N_h}\right)\right) + LeakyReLU\left(W_2\left(e_h \odot e_{N_h}\right)\right) \tag{14}$$

where $W_1, W_2 \epsilon R^{e\times e}$ are trainable weight matrices, and ⊙ denotes the element-wise product.

Based on first-layer information propagation and aggregation, high-order information propagation is realized by stacking more propagation layers. Thus, information propagation between $i$-th layer and $(i - 1)$-th's layer is formulated as:

$$e_h^{(i)} = f\left(e_h^{(i-1)}, e_{N_h}^{(i-1)}\right) \tag{15}$$

where $e_h^{(i-1)}$ is the representation of entity $h$ updated from the previous information propagation operations.

Following the information propagation and aggression, the node embedding decoding process generates the final node and relation embeddings by maximizing the likelihood of true triplets in the heterogeneous knowledge network. We then model entities and their relationships to distinct spaces (i.e. entity space and relationship spaces) and optimize their embeddings according to the translation principle. Given an existing triplet $(h, r, t)$ with $e_h$ and $e_t$ obtained after information aggregation, we project $e_h$ and $e_t$ into the relation $r$'s space, and compute the likelihood score (Lin et al., 2015):

$$g(h, r, t) = \left|\left|W_r e_h + e_r - W_r e_t\right|\right|_2^2 \tag{16}$$

where $W_r \in R^{dr\times de}$ represents the transformation matrix of relation $r$ that projects entity representations from the $de$-dimensional entity space into $dr$-dimensional relation space. In a well-trained knowledge network, the differences between likelihood scores for valid triplets and invalid triplets are maximized. We compute the likelihood scores of valid and invalid triplets to minimize the pairwise ranking loss (Lin et al., 2015):

$$L_G = \sum_{(h,r,t,t')\in T} - \ln\sigma(g(h, r, t') - g(h, r, t)) \tag{17}$$

where $\{(h, r, t) | (h, r, t) \in G_j, (h, r, t') \notin G_j\}$, $(h, r, t')$ is an invalid triplet constructed by randomly replacing an entity in a valid triplet, and $\sigma(\cdot)$ is the sigmoid function.

To summarize, in both global knowledge networks and subnetworks, information propagates attentively along the relations to retain information conducive to embedding expressiveness and filter out noise. The embedding decoding process then generates the entity global embedding and local embeddings. Attentive fusion of entity global embedding and local embeddings enriches the node embeddings, adapts to the relation evolutions, and accounts for the information timeliness in healthcare consultation success prediction. The final dynamic entity embeddings become implicit knowledge input into the multimodal data fusion and prediction layer.

**Multimodal Data Fusion and Prediction Layer:** The design of this layer aims to fuse the explicit and implicit knowledge extracted above. Fusing the dialogue embeddings and heterogeneous knowledge network embeddings is vital for addressing the data sparsity and incompleteness issue for online healthcare consultation in the virtual context. To fully exploit the interactions between multimodal features, this layer devises a new cross-attention mixed projection approach, namely cross-MPM. Representation distributions of different modalities are usually diverse and call for different types of approximation kernels for feature projection, i.e. homogeneous and inhomogeneous kernels. Figure 5 provides an overview of the patient and doctor representations learned from different modalities. The representation distributions vary among different modalities and the boundaries between different groups present different patterns. For knowledge networks, there exist soft boundaries between the patients' representations and doctors' representations. There also exist soft boundaries between the successful consultations and the failed consultations in the offline knowledge network. However, no clear boundaries exist between the successful consultations and the failed consultations in online patient-doctor dialogues. The representation distributions learned from patient-doctor dialogues from different cohorts are intertwined, indicating the insufficiency of simply using explicit knowledge to predict consultation outcomes. Consequently, the three disparate representation distributions learned from explicit and implicit knowledge encompass boundaries that require homogeneous and inhomogeneous kernels for feature projection in multimodal data fusion.

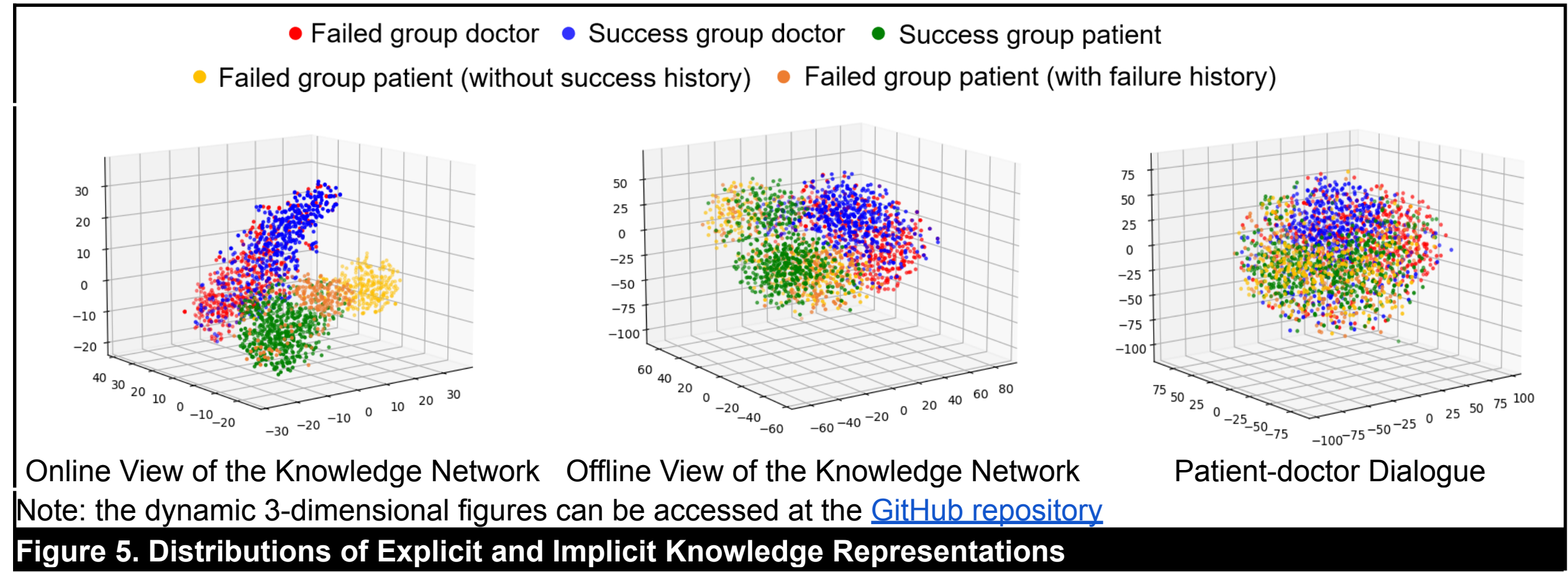

Note: the dynamic 3-dimensional figures can be accessed at the GitHub repository

**Figure 5. Distributions of Explicit and Implicit Knowledge Representations**

Conventional pooling techniques for multimodal data fusion that rely on a single projection technique with a unique kernel function may fail to capture all the necessary information in both the homogeneous and inhomogeneous distributed representations (Zanaty & Afifi, 2011; Zhang et al., 2021). Cross-MPM addresses this limitation by unifying multiple low-dimension feature maps, i.e. random Maclaurin (Kar & Karnick, 2012) and tensor sketch (Pham & Pagh, 2013), to accommodate differently distributed modality representations and project feature interactions into low-dimensional space. This mixed projection approach also helps reduce the bilinear feature dimensionality (262K) to low-dimensional features (8K). Cross-MPM allows for flexible input size and does not rely on a substantial latent dimension to deliver optimal performance, because of the attention-based cross-modal and cross-entity interaction and fusion mechanism, which reduces the dimensionality in computation and preserves important modality information.

Let $E = \{e_{mod,\,ent} | mod \in \{dialogue,\ online,\ offline\}, ent \in \{patient,\ doctor,\ hospital,\ disease\}\}$ be the multimodal representations learned from the consultation dialogues, online and offline clinical traces for different stakeholders. We use a grouping mechanism to calculate each pair of different modality representations in $E$. Taking the representation pair $online, patient$ (i.e., $e_{online, patient}$) and $offline, doctor$ (i.e., $e_{offline, doctor}$) as an example, their cross-modal and cross-entity fusion using mixed projection matrices is computed as follows:

$$v' = \frac{1}{\sqrt{d}}\left(\sum_{i=1}^{c} w_1(i)e_{online, patient}\right) \odot \left(\sum_{i=1}^{c} w_2(i)e_{offline, doctor}\right) \tag{18}$$

$$v'' = FFT^{-1}\left(FFT\left(\sum_{i=1}^{c} s_1(i)e_{online, patient}w^{H_1(i)}\right) \odot FFT\left(\sum_{j=1}^{c} s_2(j)e_{offline, doctor}w^{H_2(j)}\right)\right) \tag{19}$$

$$v = sign(v')\sqrt{|v'|} + sign(v'')\sqrt{|v''|} \tag{20}$$

where $v \in R^{d}$ is the fusion result, $w_1$, $w_2$, $s_1$, $s_2$ are independent and generated randomly using Rademacher vectors: $[c] \rightarrow \{+1, -1\}$, and $w^{H_1(i)}$, $w^{H_1(i)}$ are count sketch hashing operations with $H_1(i)$, $H_2(j)$ uniformly drawn from $\{1,..., d\}$. Applying fusion operation to every pair of representations in $E$ can fully explore the interactions between features of different stakeholders and modalities. Supposedly $V = \{v_1,..., v_{|E|(|E|-1)/2}\}$ is a set of sub-group fusion outputs, Cross-MPM then combines these fusion results with attentional weights as follows:

$$\alpha_z = \frac{exp\left(LeakyReLU\left(W_{cross} \cdot v_z\right)\right)}{\Sigma_{v_z \in V} exp\left(LeakyReLU\left(W_{cross} \cdot v_z\right)\right)} \tag{21}$$

$$v_{mix} = \Sigma_{v_z \in V} \alpha_z v_z \tag{22}$$

where $\alpha_z$ is the attention weight, $W_{cross} \in R^{d \times 1}$ is the parameter matrix, and $d$ is the output vector dimensionality. The algorithm of cross-MPM is shown in the GitHub repository.

With the fusion output $v_{mix}$, we use a feed-forward neural network with three layers and LeakyReLU activation to obtain the consultation success prediction result $\widetilde{y}_s$. The prediction loss function is:

$$L_{re} = -\frac{1}{N}\left(\sum_{s \in R^{+}} log\sigma\left(\widetilde{y}_s\right) + \sum_{s \in R^{-}} log\left(1 - \sigma\left(\widetilde{y}_s\right)\right)\right) \tag{23}$$

where $N$ denotes the number of training instances. $R^{+}$ and $R^{-}$ denote the set of positive and negative consultation instances. The complete loss function for DyKoNeM is composed of the prediction loss and knowledge network training loss:

$$L_{overall} = L_{G} + L_{re} + \lambda||\Theta||^{2} \tag{24}$$

where $\Theta$ denotes model trainable parameters, and $L_{G}$ is computed according to Eq. 17. We optimize $L_{G}$ and $L_{re}$ alternatively in the optimization of $L_{overall}$.

In sum, our proposed framework presents two key novelties. First, we propose a novel dynamic knowledge network structure with an attribute encoder that extends the existing static knowledge networks considering solely entities and their relationships. Our study takes into account the attributes associated with each entity. These attributes disclose more implicit information between stakeholders in the online and offline healthcare systems (i.e., online and offline view of the knowledge network). The inclusion of entity attributes is crucial for generating accurate knowledge networks and representing entities effectively. Furthermore, the attributes of these entities exhibit cross-influence as information propagates through the knowledge networks. For instance, patient attributes are valuable for creating a comprehensive patient representation and play a critical role in downstream healthcare consultation success prediction. Another example is the hospital tiers and the affiliation between the doctor and the hospital. These factors reflect the diseases that the doctor specializes in, imply the patient's expectations of the doctor, and provide valuable insights for healthcare consultation success prediction. These complex interactions have a significant impact on the generation of entity representations, making it inadequate to simply concatenate them with the representations generated by the vanilla knowledge graph attention network (KGAT) approach. Consequently, we aim to cohesively learn them together with the knowledge network, so that the complex interconnected structures among these attributes can be fruitfully captured and implicit knowledge in the knowledge network can be further exploited.

Therefore, we propose a novel attribute encoder that effectively captures the intricate attributes associated with the entities. By integrating this attribute encoder into the learning and generation process of the knowledge network, we not only gain insights into the entities and their connections but also facilitate a deeper understanding and representation of the complex attributes of the entities. Moreover, it enables us to examine the impact of the interactions among these attributes on the entire knowledge network. This innovation holds significant relevance. In many problem domains and applications of knowledge networks, entities are accompanied by attributes that contain crucial and intricate information. Incorporating these attributes when generating knowledge networks has significant potential to enhance the performance of downstream tasks.

In addition, static knowledge network models (e.g. KGAT) treat "knowledge" observed in different periods equally. However, knowledge in the health domain is time-sensitive and has disparate predictive power for future events. Patients' medical needs also evolve in different health conditions and stages. In our research design, we propose a temporal knowledge aggregation layer to address this issue. This layer is designed to focus on informative knowledge for the current prediction, allowing us to prioritize relevant information.

Second, we propose a cross-attention mixed projection approach to alleviate the information loss during multimodal data fusion. Our method seamlessly and attentively integrates multiple approximation kernels for feature projection, allowing them to complement one another and compensate for information loss incurred by individual algorithms. Consequently, the original multimodal data's information can be better preserved, enhancing its performance for downstream prediction tasks.

## EVALUATION

### Comparison with Existing Healthcare Consultation Success Prediction Studies

Because the online healthcare consultation success prediction in the virtual context is a new research problem, no existing method has been developed for this problem; a closely related problem is in-person healthcare consultation success prediction. The existing research on predicting healthcare consultation success is predominantly centered around in-person healthcare settings (Table 4), employing machine learning predictive models and leveraging the extensive data present in health IT systems, such as patients' complete medical history (Giesemann et al., 2023; Pradier et al., 2020), comprehensive socio-economic demographic information (Bennemann et al., 2022; Maskew et al., 2022; Mütze et al., 2022; Pedersen et al., 2019; Pradier et al., 2020; Ramachandran et al., 2020), disease progression (Mütze et al., 2022), specific laboratory test results (Maskew et al., 2022; Ramachandran et al., 2020), and patient survey data (Bennemann et al., 2022; Mütze et al., 2022). Notably, the majority of current healthcare consultation success studies are tailored to specific patient cohorts focused on the management of particular medical conditions (Table 4).

**Table 4. Representative Healthcare Consultation Success Prediction Studies**

| Literature | Patient Cohorts | Delivery Model | Methods | Features |
|---|---|---|---|---|
| Ramachandran et al. (2020) | HIV | In-person | Gradient boosting decision trees | Demographics, diagnoses, location, lab tests, medical visits, providers seen |
| Pedersen et al. (2019) | Chronic disease | In-person | Random forest | Demographics, patient behavior |
| Bennemann et al. (2022) | Mental disorder | In-person | Random forest, nearest neighbor, GLM stacking | Healthcare outcome expectations questionnaire results, demographics |
| Mütze et al. (2022) | Mental disorder | In-person | LASSO | Demographics, disease history, symptom severity, motivational factors |
| Maskew et al. (2022) | HIV | In-person | AdaBoost | Demographic, clinical, behavioral, and laboratory |
| Pradier et al. (2020) | Depression | In-person | Random Forest | Medical history, demographics |
| Giesemann et al. (2023) | Mental disorder | In-person | Resampling methods, 20 ML Algorithms | Clinical records |

We first compare with state-of-the-art studies in healthcare consultation success prediction. They include Pedersen et al. (2019), Ramachandran et al. (2020), Pradier et al. (2020), Maskew et al. (2022), Bennemann et al. (2022), Mütze et al. (2022), and Giesemann et al. (2023). The hyperparameters are fine-tuned using large-scale experiments to reflect their best capability in our dataset. Comparison with these studies validates our model in the same context as the existing methods.

As healthcare consultation success prediction is a classification problem, we adopt F1-score, precision, and recall as the evaluation metrics. The best model should have the highest F1-score. We use 80% of our data as the training set, 10% as the validation set, and 10% as the test set. The following performances are the mean of 10 experimental runs. For the deep learning-based models, we also report the standard error of the performances to show their statistical significance. As reported in Table 5, compared to the leading healthcare consultation success prediction study (Bennemann et al., 2022), our model improves the F1-score by 0.1183. These studies mostly use traditional machine learning approaches, which fall short in capturing implicit knowledge from the dynamic knowledge network and integrating explicit and implicit knowledge from multimodal data.

**Table 5. Comparison with State-of-the-art healthcare Consultation Success Prediction Studies**

| Model | | F1 | Precision | Recall |
|---|---|---|---|---|
| **DyKoNeM (Ours)** | | **0.7822 ± 0.0009** | **0.8950 ± 0.0044** | **0.6949 ± 0.0038** |
| Pedersen et al. (2019) | | 0.5618 | 0.6757 | 0.4808 |
| Ramachandran et al. (2020) | | 0.5686 | 0.7404 | 0.4615 |
| Pradier et al. (2020) | | 0.6241 ± 0.0046 | 0.7867 ± 0.0058 | 0.5175 ± 0.0052 |
| Maskew et al. (2022) | | 0.4931 | 0.5788 | 0.4295 |
| Bennemann et al. (2022) | | 0.6580 ± 0.0017 | 0.7320 ± 0.0030 | 0.5976 ± 0.0020 |
| Mütze et al. (2022) | | 0.2747 | 0.5534 | 0.1827 |
| Giesemann et al. (2023) | | 0.6150 ± 0.0049 | 0.7644 ± 0.0046 | 0.5146 ± 0.0055 |
| LLM | BERT (fine-tune with labeled data) | 0.7405 ± 0.0010 | 0.8192 ± 0.0074 | 0.6747 ± 0.0035 |
| | ChatGPT-4o* (one-shot learning) | 0.0758 | 0.0732 | 0.0785 |

Note: * For ChatGPT-4o, we use the following prompt: "*Here is a passage of a patient-doctor dialogue from an online consultation. Based on this dialogue, determine whether the consultation is successful. If the consultation is not successful, output 1. If the consultation is successful, output 0. Do not output anything else.*"

With the rapid development in the field of natural language processing, LLMs have gained the ability to conduct text data-based predictions. To demonstrate that our predictions surpass the capabilities of the state-of-the-art LLM, we compare our framework with the prediction results of BERT (fine-tune with labeled data) and ChatGPT-4o (one-shot learning). Since LLMs cannot handle knowledge networks, only patient-doctor dialogues are used as input for the LLM. The results are reported in Table 5. The results indicate that predicting the success of healthcare consultations is challenging. Although powerful, LLMs still fall short in handling nuanced network features from knowledge networks that collectively inform the prediction of success in online healthcare consultations. Without such implicit knowledge, BERT's performance is significantly poorer than that of our proposed framework. Moreover, the prediction results are inadequate for ChatGPT-4o using a prompt and only patient-doctor dialogues for one-shot learning. Therefore, our proposed framework cannot be simply replaced by LLMs.

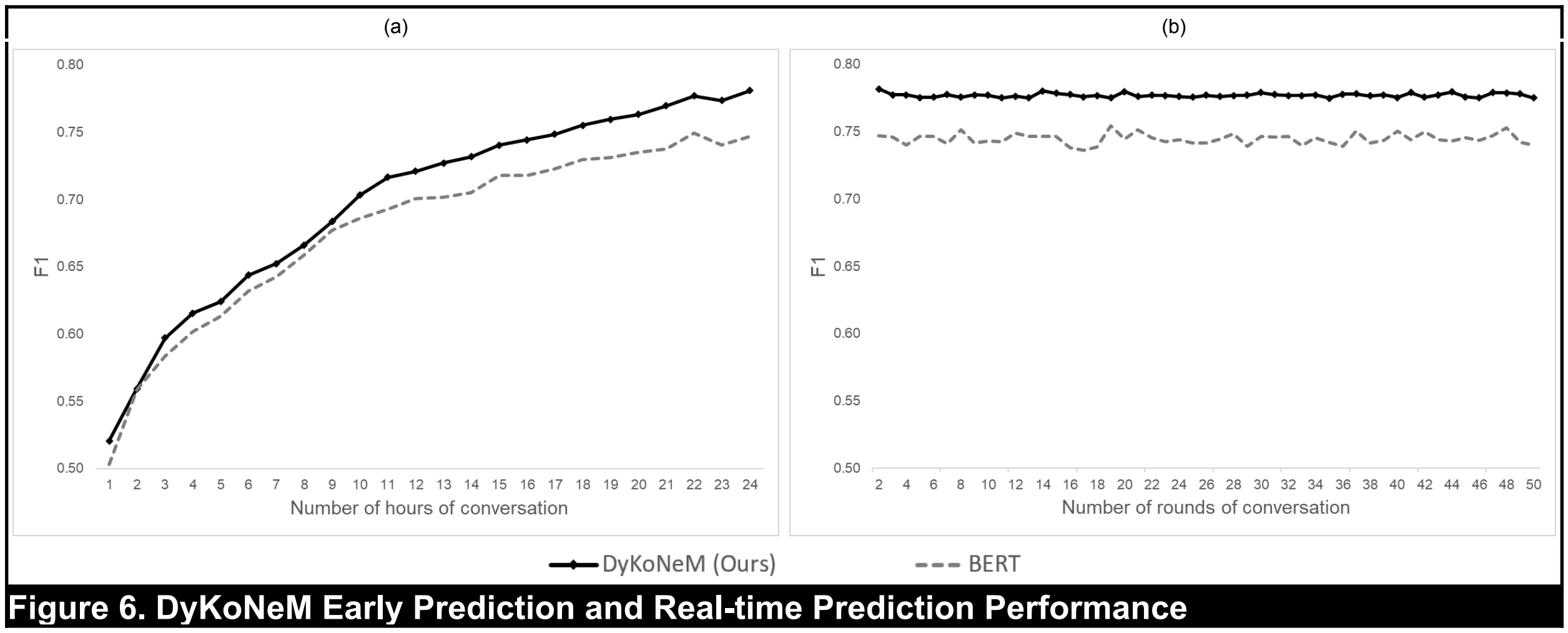

**Figure 6. DyKoNeM Early Prediction and Real-time Prediction Performance**

As previously mentioned, accurately and promptly predicting the success of healthcare consultations allows online healthcare platforms to take proactive steps to address patients' concerns and needs, potentially improving patient retention rates. We also test our framework's early prediction and real-time prediction capabilities (Figure 6). The evaluations are conducted in two ways. (1) Based on the number of hours of conversation: instead of using the entire patient-doctor communication process as input (24 hours in total), we generate prediction results every hour. These results are presented in Figure 6(a). The findings indicate that performance improves as more hours of data are used; our proposed framework consistently outperforms the best-performing baseline in Table 5 (i.e., BERT). It also achieves early prediction: with 10 hours of data, our framework's performance surpasses the best-performing baseline that uses 24 hours of data. Therefore, by combining this with the reasons for failed consultations (see Table 18), the platform can intervene before the consultation process is closed and before patient loss occurs.

(2) Based on the number of rounds of conversation: Instead of using the entire patient-doctor dialogues, we generate a prediction each round the patient and doctor exchange information. The results are reported in Figure 6(b). The performance of our proposed framework consistently surpasses the best-performing baseline in Table 5 (i.e., BERT). The results demonstrate that our framework can successfully achieve real-time prediction: since a new prediction is generated with each exchange of information between the patient and doctor, these real-time prediction results can be provided to both doctors and the platform. It enables real-time monitoring of the patient's status, thereby reducing the number of failed consultations.

## Robustness Analyses

**Effectiveness of Each Design Component:** Our method is composed of multiple critical design components. We first delineate the individual effect of each component in Table 6. The patient-doctor dialogues are the most essential information of the online health consultation and directly affect patient consultation success. Model A only considers the patient-doctor dialogues and reaches a base F1-score of 0.6186. Our virtual health collaborator partners with hospitals across China and has access to its registered patients' offline appointment information. Using patients' offline digital traces throughout their healthcare journeys, we construct the offline view of the knowledge network. Model B considers the added value of this offline component to patient-doctor dialogues in consultation success prediction. The offline component boosts the base F1-score to 0.7170. This suggests that the integration of offline health digital traces significantly improves the prediction for online health platforms. This multi-channel (i.e., online and offline) patient information integration via a healthcare platform can be an exemplary business model for many other countries and regions. In addition to the offline data, our virtual health partner collects patients' online consultation traces. Similarly, we construct the online view of the knowledge network. Model C shows its added value to patient-doctor dialogues. This online view of the knowledge network (F1 = 0.7519) is shown to be more effective than the offline view of the knowledge network. Moreover, as described in the Research Design section, we develop a dynamic online view of the knowledge network. An alternative design is a static network, where the network does not change over time. Model D shows the effect when using the static network (F1 = 0.7566), which is proven to be inferior to our dynamic network design. Considering all the design components, our framework, DyKoNeM, reaches the highest F1-score of 0.7822.

**Table 6. Individual Effect of Each Design Component**

| Model | F1 | Precision | Recall |
|---|---|---|---|
| Model A: Dialogue model | 0.6186 ± 0.0008 | 0.6842 ± 0.0045 | 0.5647 ± 0.0031 |
| Model B: Dialogue + offline view of the network | 0.7170 ± 0.0017 | 0.7453 ± 0.0123 | 0.6929 ± 0.0078 |
| Model C: Dialogue + dynamic online view of the network | 0.7519 ± 0.0017 | 0.8466 ± 0.0076 | 0.6766 ± 0.0028 |
| Model D: Dialogue + static online view of the network + offline view of the network | 0.7566 ± 0.0008 | 0.8593 ± 0.0057 | 0.6761 ± 0.0027 |
| **DyKoNeM: Dialogue + dynamic online view of the network + offline view of the network** | **0.7822 ± 0.0009** | **0.8950 ± 0.0044** | **0.6949 ± 0.0038** |

**Robustness Analysis of the Dialogue Embedding Learning Layer:** This layer of our framework encodes explicit patient-doctor dialogues, where temporal information is considered (Eq. 5). We remove this temporal information to test its effectiveness. As reported in Table 7, removing it reduces model performance. Therefore, including it in our design is appropriate.

**Table 7. Effectiveness of Temporal Information in the Dialogue Embedding Learning Layer**

| Model | F1 | Precision | Recall |
|---|---|---|---|
| **DyKoNeM (Ours: with temporal information)** | **0.7822 ± 0.0009** | **0.8950 ± 0.0044** | **0.6949 ± 0.0038** |
| No temporal information | 0.7701 ± 0.0007 | 0.8882 ± 0.0069 | 0.6803 ± 0.0038 |

**Robustness Analysis of the Knowledge Network Construction and Attribute Encoding Layer:** We test alternative designs to validate the effectiveness of the attribute encoding layer in the knowledge network of our framework. As reported in Table 8. We first remove all attributes of the entities. The F1 drops to 0.7593. These attributes encode critical information of doctors and hospitals that are transparent to patients when starting the consultation process and selecting doctors. The transparency of this attribute is unique to China and our research context, potentially offering implications for healthcare service delivery in other regions. We also combine the online network and offline views of the knowledge network in a single network.

The F1 reduces to 0.7594. Next, we employ the sum pooling and concatenation for attribute encoding successively. The F1 drops to 0.7758 and 0.7704, respectively. These results suggest that our multi-view knowledge network design and attribute encoding of our knowledge network are effective.

**Table 8. Robustness Analyses of the Dialogue Embedding Learning Layer**

| **Model** | **F1** | **Precision** | **Recall** |
|---|---|---|---|
| **DyKoNeM (Ours)** | **0.7822 ± 0.0009** | **0.8950 ± 0.0044** | **0.6949 ± 0.0038** |
| Remove attributes | 0.7593 ± 0.0013 | 0.8643 ± 0.0064 | 0.6774 ± 0.0029 |
| Combine online and offline views in one network | 0.7594 ± 0.0005 | 0.8582 ± 0.0030 | 0.6811 ± 0.0019 |
| Attribute encoding by sum pooling | 0.7758 ± 0.0007 | 0.8659 ± 0.0054 | 0.7030 ± 0.0043 |
| Attribute encoding by concatenation | 0.7704 ± 0.0008 | 0.8579 ± 0.0052 | 0.6994 ± 0.0033 |

**Robustness Analysis of the Dynamic and Attentive Network Embedding Learning Layer:** This layer of our framework includes a dynamic knowledge network design to capture the evolving relations of patients, doctors, diseases, and hospitals. To construct such a dynamic knowledge network, we use a sliding window of four weeks to develop a time series of sub-networks. To validate that this four-week window is optimal, we test the effect of the sliding window length on the prediction performance, reported in Table 9. Moreover, our model opts for a six-month window to construct the sub-networks. Table 10 shows the influence of the observation window size on the healthcare consultation success prediction performance. The results indicate that the prediction performance improves with an extended observation window.

**Table 9. The Effect of Sliding Window of the Dynamic Network**

| **Sliding Window** | **F1** | **Precision** | **Recall** |
|---|---|---|---|
| Two weeks | 0.7606 ± 0.0014 | 0.8956 ± 0.0051 | 0.6612 ± 0.0021 |
| **Four weeks** | **0.7822 ± 0.0009** | **0.8950 ± 0.0044** | **0.6949 ± 0.0038** |
| Six weeks | 0.7628 ± 0.0008 | 0.8743 ± 0.0080 | 0.6772 ± 0.0051 |
| Eight weeks | 0.7592 ± 0.0016 | 0.8455 ± 0.0047 | 0.6894 ± 0.0027 |

**Table 10. The Effect of the Observation Window of the Sub-networks**

| **Observation Window** | **F1** | **Precision** | **Recall** |
|---|---|---|---|
| One month | 0.7642 ± 0.0008 | 0.8648 ± 0.0081 | 0.6854 ± 0.0056 |
| Two months | 0.7645 ± 0.0013 | 0.8741 ± 0.0058 | 0.6796 ± 0.0036 |
| Three months | 0.7682 ± 0.0009 | 0.8826 ± 0.0060 | 0.6804 ± 0.0032 |
| **Six months** | **0.7822 ± 0.0009** | **0.8950 ± 0.0044** | **0.6949 ± 0.0038** |

In our dynamic network, we include the patient entity, doctor entity, hospital entity, and disease entity. Each entity type contributes to an essential part of the knowledge network. To verify the effectiveness of each entity type, we conduct an ablation study of the dynamic network, that is, removing one entity type at a time. Table 11 reports the ablation study results. Removing any entity significantly hampers the prediction performance of the dynamic network. Therefore, our design of the dynamic network is the most appropriate.

**Table 11. Ablation Studies of the Dynamic Network**

| Model | F1 | Precision | Recall |
|---|---|---|---|
| **DyKoNeM (Ours)** | **0.7822 ± 0.0009** | **0.8950 ± 0.0044** | **0.6949 ± 0.0038** |
| Remove patient entities | 0.6860 ± 0.0008 | 0.8364 ± 0.0080 | 0.5819 ± 0.0033 |
| Remove doctor entities | 0.7110 ± 0.0022 | 0.7613 ± 0.0132 | 0.6697 ± 0.0095 |
| Remove hospital entities | 0.7568 ± 0.0010 | 0.8342 ± 0.0055 | 0.6929 ± 0.0046 |
| Remove disease entities | 0.7526 ± 0.0006 | 0.8464 ± 0.0046 | 0.6777 ± 0.0031 |

**Robustness Analysis of the Multimodal Data Fusion and Prediction Layer:** This layer of our framework is our proposed fusion method. We first remove this fusion method to test its effectiveness. Without the fusion method, we can directly feed the concatenated network representations from previous layers into a machine learning or deep learning classifier to predict healthcare consultation success. For machine learning classifiers, we test decision trees (DT), logistic regression (LR), random forest (RF), support vector machine (SVM), and Linear Discriminant Analysis (LDA). For deep learning classifiers, we test neural network (NN), convolutional neural network (CNN), recurrent neural network (RNN), long short-term memory (LSTM), and Transformer.

**Table 12. Effectiveness of the Fusion Method**

| Model | F1 | Precision | Recall |
|---|---|---|---|
| **DyKoNeM (Ours: with fusion)** | **0.7822 ± 0.0009** | **0.8950 ± 0.0044** | **0.6949 ± 0.0038** |
| No fusion with DT | 0.6182 ± 0.0022 | 0.5957 ± 0.0029 | 0.6426 ± 0.0023 |
| No fusion with LR | 0.6151 | 0.7229 | 0.5353 |
| No fusion with RF | 0.6606 ± 0.0029 | 0.6750 ± 0.0036 | 0.6470 ± 0.0040 |
| No fusion with SVM | 0.5579 | 0.5059 | 0.6218 |
| No fusion with LDA | 0.5446 | 0.6149 | 0.4888 |
| No fusion with NN | 0.7329 ± 0.0009 | 0.8118 ± 0.0046 | 0.6683 ± 0.0029 |
| No fusion with CNN | 0.7351 ± 0.0011 | 0.8564 ± 0.0056 | 0.6442 ± 0.0029 |
| No fusion with RNN | 0.7153 ± 0.0012 | 0.8441 ± 0.0081 | 0.6212 ± 0.0033 |
| No fusion with LSTM | 0.7070 ± 0.0019 | 0.8196 ± 0.0077 | 0.6221 ± 0.0036 |
| No fusion with Transformer | 0.7410 ± 0.0010 | 0.8311 ± 0.0074 | 0.6691 ± 0.0035 |

Table 12 reports the results when we use different machine and deep learning classifiers with our network representations from previous layers as their inputs. When we use machine learning classifiers coupled with our network representations, compared to the best-performing traditional machine learning model (RF), our fusion method improves F1-score by 0.1216. When we use deep learning classifiers coupled with our network representations, compared to the best-performing deep learning structure (Transformer), our fusion method increases F1-score by 0.0412. Such superior performance is also robust in precision and recall.

There are many other fusion strategies. However, many of them are tailored to certain specific data modalities, such as image or audio, which imposes strict restrictions on the input, making them inapplicable to our study. In Table 13, we test the effect of seven alternative fusion strategies that are the most relevant to us. The results indicate our fusion design achieves the best performance, thus optimal.

Our fusion method contains a grouping mechanism (Eq. 18 - Eq. 20). We test its effectiveness in Table 14. The result indicates that without it, the model performance drops while the latent feature dimension increases. Therefore, this design is effective in our research design.

**Table 13. The Effect of Alternative Fusion Strategies**

| Fusion Strategy | F1 | Precision | Recall |
|---|---|---|---|
| **DyKoNeM (Our fusion method)** | **0.7822 ± 0.0009** | **0.8950 ± 0.0044** | **0.6949 ± 0.0038** |
| Concatenation fusion | 0.7426 ± 0.0003 | 0.8540 ± 0.0054 | 0.6572 ± 0.0029 |
| Average fusion | 0.7472 ± 0.0005 | 0.8417 ± 0.0026 | 0.6718 ± 0.0015 |
| Tensor sketch (Pham & Pagh, 2013) | 0.7595 ± 0.0004 | 0.8617 ± 0.0077 | 0.6796 ± 0.0050 |
| Random Maclaurin (Kar & Karnick, 2012) | 0.7569 ± 0.0012 | 0.8340 ± 0.0109 | 0.6944 ± 0.0072 |
| Tensor fusion network (Zadeh et al., 2017) | 0.7593 ± 0.0014 | 0.8618 ± 0.0069 | 0.6790 ± 0.0043 |
| Low-rank multimodal fusion (Liu et al., 2018) | 0.7487 ± 0.0012 | 0.8493 ± 0.0073 | 0.6700 ± 0.0050 |
| Gated multimodal (Arevalo et al., 2017) | 0.7584 ± 0.0005 | 0.8606 ± 0.0047 | 0.6782 ± 0.0030 |

**Table 14. Effectiveness of Grouping**

| Model | Latent feature dimension / parameter scale | F1 | Precision | Recall |
|---|---|---|---|---|
| **DyKoNeM (Ours: with grouping)** | **16,897,056** | **0.7822 ± 0.0009** | **0.8950 ± 0.0044** | **0.6949 ± 0.0038** |
| No grouping | 4,224,352 | 0.7705 ± 0.0017 | 0.8672 ± 0.0052 | 0.6934 ± 0.0033 |

## Patient Group Analyses

In the previous analyses, we use all patients in our dataset to predict healthcare consultation success. In reality, different hospital departments have significantly different consultation success rates, resulting from the distinct complexity and severity of diseases. To show that our model is robust in predicting healthcare consultation success for patients with various diseases, we perform a heterogeneity analysis of diseases.

**Table 15. Analysis of Heterogeneity of Diseases**

| Disease | Failure Rate | % Cases | F1 | Precision | Recall |
|---|---|---|---|---|---|
| All | 5.18% | 100.00% | 0.7822 ± 0.0009 | 0.8950 ± 0.0044 | 0.6949 ± 0.0038 |
| Dermatology | 5.66% | 12.77% | 0.7587 ± 0.0043 | 0.9061 ± 0.0108 | 0.6545 ± 0.0116 |
| Gynecology | 6.73% | 8.59% | 0.7767 ± 0.0016 | 0.9401 ± 0.0138 | 0.6630 ± 0.0054 |
| Dentistry | 7.15% | 4.79% | 0.9050 ± 0.0015 | 0.9045 ± 0.0133 | 0.9095 ± 0.0150 |
| Endocrinology | 6.07% | 4.56% | 0.8677 ± 0.0019 | 0.9952 ± 0.0048 | 0.7692 ± 0.0000 |
| Ear, nose and throat | 4.60% | 3.67% | 0.8016 ± 0.0039 | 0.9273 ± 0.0121 | 0.7071 ± 0.0071 |
| Reproductive medicine | 4.29% | 3.54% | 0.8288 ± 0.0023 | 0.9364 ± 0.0139 | 0.7462 ± 0.0118 |
| Chinese medicine | 4.76% | 2.86% | 0.7696 ± 0.0058 | 0.7968 ± 0.0055 | 0.7455 ± 0.0121 |
| Urology | 5.53% | 2.74% | 0.8116 ± 0.0041 | 0.7928 ± 0.0094 | 0.8333 ± 0.0111 |
| Orthopedics | 6.99% | 2.71% | 0.8767 ± 0.0034 | 0.9645 ± 0.0078 | 0.8043 ± 0.0072 |
| Ophthalmology | 4.93% | 2.66% | 0.8515 ± 0.0062 | 0.9111 ± 0.0148 | 0.8000 ± 0.0000 |
| Hematology | 4.96% | 2.55% | 0.9004 ± 0.0016 | 0.9970 ± 0.0030 | 0.8211 ± 0.0035 |
| Psychology | 11.72% | 2.22% | 0.8518 ± 0.0022 | 0.8884 ± 0.0047 | 0.8182 ± 0.0000 |
| General surgery | 4.88% | 2.20% | 0.8000 ± 0.0000 | 1.0000 ± 0.0000 | 0.6667 ± 0.0000 |
| General practice | 7.85% | 1.96% | 0.8451 ± 0.0020 | 0.8747 ± 0.0094 | 0.8200 ± 0.0133 |
| Neonatology | 7.86% | 1.74% | 0.9787 ± 0.0000 | 1.0000 ± 0.0000 | 0.9583 ± 0.0000 |
| Respiratory internal medicine | 5.80% | 1.25% | 0.8514 ± 0.0057 | 0.8464 ± 0.0107 | 0.8571 ± 0.0000 |
| Neurology | 7.05% | 1.23% | 0.8779 ± 0.0085 | 1.0000 ± 0.0000 | 0.7833 ± 0.0136 |
| Colorectal surgery | 6.68% | 1.10% | 0.8571 ± 0.0000 | 1.0000 ± 0.0000 | 0.7500 ± 0.0000 |
| Infectious disease | 5.18% | 1.06% | 0.7500 ± 0.0000 | 1.0000 ± 0.0000 | 0.6000 ± 0.0000 |
| Cardiology | 6.64% | 1.00% | 0.9359 ± 0.0052 | 0.8800 ± 0.0089 | 1.0000 ± 0.0000 |
| Andrology | 14.07% | 0.63% | 0.9677 ± 0.0116 | 0.9618 ± 0.0207 | 0.9778 ± 0.0148 |
| Pediatric respiratory | 1.96% | 0.54% | 1.0000 ± 0.0000 | 1.0000 ± 0.0000 | 1.0000 ± 0.0000 |
| Neurosurgery | 4.41% | 0.53% | 0.7607 ± 0.0107 | 0.7750 ± 0.0250 | 0.7500 ± 0.0000 |
| Pediatric surgery | 2.29% | 0.48% | 0.9333 ± 0.0444 | 0.9000 ± 0.0667 | 1.0000 ± 0.0000 |
| General surgery | 5.69% | 0.45% | 0.8000 ± 0.0000 | 1.0000 ± 0.0000 | 0.6667 ± 0.0000 |

Table 15 shows the consultation failure rates of the most common diseases of our business partner as well as the F1-score, precision, and recall of our framework in each of these diseases. The results suggest our framework is robust and reaches consistently high performance in predicting healthcare consultation success. Additionally, there is heterogeneity among different diseases. Specifically, our framework performs the best for pediatric respiratory, neonatology, and andrology. Andrology accounts for the highest consultation failure rate. Larger number of failure instances benefits the predictive model training. Pediatric respiratory has a low consultation failure rate, indicating that DyKoNeM can also accurately identify the failure instances when training samples from the same disease are limited. Neonatology treats newborns up to 28 days old and usually only minor health concerns are asked in asynchronous health consultations. Cases of dermatology, gynecology, dentistry, endocrinology, ENT (Ear, nose and throat), Chinese medicine, urology, orthopedics, ophthalmology, and hematology, psychology, general surgery, general practice, respiratory internal medicine, neurology, colorectal surgery, infectious disease, cardiology, neurosurgery, pediatric surgery, general surgery, where our model consistently performs high (mostly higher than the performance for all patients), constitute the majority of the online health consultations (67.83%).

We break down the patients into various gender and age groups and test our framework's effectiveness. Tables 16 (Gender Group and Age Group) show that our framework achieves consistently high performance in all gender and age groups.

**Table 16. Analysis of Patient Gender / Age / Location Groups**

| Patient Group | | Failure Rate | % Cases | F1 | Precision | Recall |
|---|---|---|---|---|---|---|
| All | | 5.18% | 100.00% | 0.7822 ± 0.0009 | 0.8950 ± 0.0044 | 0.6949 ± 0.0038 |
| Gender Group | Male | 6.30% | 34.68% | 0.8018 ± 0.0009 | 0.8635 ± 0.0023 | 0.7484 ± 0.0025 |
| | Female | 5.69% | 65.32% | 0.7482 ± 0.0011 | 0.8600 ± 0.0091 | 0.6631 ± 0.0062 |
| Age Group | 0-2 | 3.85% | 12.08% | 0.8370 ± 0.0016 | 0.9182 ± 0.0213 | 0.7746 ± 0.0150 |
| | 3-6 | 4.57% | 8.75% | 0.8137 ± 0.0014 | 0.9407 ± 0.0038 | 0.7170 ± 0.0000 |
| | 7-12 | 4.95% | 6.32% | 0.8476 ± 0.0029 | 0.9633 ± 0.0100 | 0.7575 ± 0.0053 |
| | 13-17 | 7.69% | 2.08% | 0.8889 ± 0.0000 | 0.9412 ± 0.0000 | 0.8421 ± 0.0000 |
| | 18-25 | 9.30% | 10.73% | 0.8312 ± 0.0017 | 0.8571 ± 0.0076 | 0.8079 ± 0.0076 |
| | 26-45 | 5.92% | 51.28% | 0.7160 ± 0.0009 | 0.8543 ± 0.0114 | 0.6175 ± 0.0060 |
| | 46-69 | 6.17% | 7.73% | 0.7621 ± 0.0040 | 0.8522 ± 0.0086 | 0.6897 ± 0.0046 |
| | 70+ | 5.11% | 1.03% | 1.0000 ± 0.0000 | 1.0000 ± 0.0000 | 1.0000 ± 0.0000 |
| Location Group | Local | 5.64% | 82.82% | 0.7707 ± 0.0007 | 0.8971 ± 0.0074 | 0.6762 ± 0.0048 |
| | Cross-city | 7.18% | 17.18% | 0.8264 ± 0.0015 | 0.9221 ± 0.0038 | 0.7489 ± 0.0033 |

One unique advantage of virtual health platforms over in-person healthcare services is that they allow for health consultations free of geographic constraints. While local patients (patient residence city is the same as doctor residence city) are important for the online platform, cross-city patients (patient residence city is different from doctor residence city) are strategic targets of the online platform, because cross-city patients allow the platform to expand into suburban and rural areas. This is especially true in China, where the best and majority of health resources are concentrated in a few megacities. Patients in suburban and rural areas usually do not have access to high-quality healthcare services. The virtual health platform addresses this challenge. Due to the significance of cross-city patients, we split the patients into local patients and cross-city patients and test our model in both groups. Table 16 (Location Group) reports the results. Our framework reaches consistently high performance in both groups. In particular, our model has a higher F1-score (0.8264) in cross-city patients than in all patients. Cross-city patients represent a significant and promising demographic for our business partner's future expansion.

## DISCUSSION

The global healthcare industry is facing the challenge of providing accessible and high-quality healthcare. Virtual health applications have emerged as a promising solution, but face intense competition. Numerous proactive interventions can be implemented by platforms upon a prompt and accurate prediction of the success of online healthcare consultations. This ensures the survival and long-term development of online healthcare platforms while maximizing the benefits they bring to patients. Because of the point-solution nature and complementary role of online healthcare consultations compared to in-person healthcare, patient information in the virtual context is notably sparse, fragmented, and incomplete, posing challenges to healthcare consultation success prediction. To address this issue, we develop DyKoNeM that devises both a language model to represent patient-doctor communication processes and the dynamic knowledge graph attention networks to integrate latent information from key stakeholders' digital traces in the virtual healthcare context. We evaluate the proposed framework using real-world data from one of the largest virtual health platforms in China. The experimental results show that our framework enhances the predictive power of healthcare consultation success for virtual health platforms and outperforms state-of-the-art healthcare consultation success prediction methods.

We have identified three research questions: (1) how to effectively represent explicit knowledge, (2) how to effectively represent implicit knowledge, and (3) how to integrate explicit and implicit knowledge to improve the prediction success of online healthcare consultations. To address these research questions, we first devise a new language model representation that captures the time intervals between dialogues. Next, we develop a novel knowledge network that represents multi-view perspectives, encodes and propagates attributes associated with nodes, and accounts for network dynamics. Finally, we create a novel fusion method that considers the unique data distributions inherent in various data modalities and balances the efficiency of data fusion with the creation of a comprehensive representation.

From the perspective of design science, we make two contributions. First, we propose a novel framework for predicting the success of online healthcare consultations using a dynamic multiview knowledge graph attention network and multimodal fusion. Second, as part of our framework, we propose two novel IT artifacts, which include 1) the new dynamic knowledge graph attention networks for representing the complicated relationships among various stakeholders with multiple views, and 2) the new multimodal fusion method to effectively integrate information from explicit and implicit knowledge.

The practical implications of this study are significant. Compared to related studies in this area, DyKoNeM is able to accurately predict the most failed consultations, as shown in Table 17. On our business partner's platform, our method accurately predicted 226,806 more failed consultations than the best-performing benchmark (Bennenmann et al. 2022). Predicting these failed consultations brings significant value to the platform, as various interventions can be implemented to retain patients and improve user satisfaction. The common consultation failure reasons and corresponding intervention strategies are summarized in Table 18. Furthermore, the implications of our work can yield benefits for various stakeholders. First, on the patient level, the ability to predict healthcare consultation success in virtual health platforms is a critical aspect of patient care that can greatly benefit patients with healthcare needs. Healthcare consultation success is essential to ensure that patients receive consistent and continuous care from their providers, which is crucial for managing chronic diseases and ensuring positive patient outcomes. Second, on the virtual health provider and platform level, it is important for virtual health providers to predict which patients may be at risk of leaving the service. By leveraging data and analytics, virtual health providers can identify patients who may be struggling with the technology, experiencing difficulty with their healthcare plan, or simply disengaging with the service. This allows providers to intervene early and address any concerns or needs of patients, potentially improving healthcare consultation success. Our accurate prediction results enable virtual health providers to focus on providing individualized care, leading to improved patient engagement and trust. This approach is likely to be highly effective, as it allows virtual health providers to cater to the unique requirements of each patient, increasing the likelihood of success. Third, on the population level, the ability to predict healthcare consultation success can help improve healthcare quality and outcomes. When patients stay engaged with the service and receive consistent care, their health outcomes are likely to be better. This is particularly important for patients with chronic diseases or conditions that require ongoing care and monitoring. Healthcare providers can help to ensure that patients receive the care they need, ultimately leading to better healthcare outcomes on the population level.

**Table 17. Accurately Identified Failed Consultations Compared to State-of-the-art Methods**

| **Model** | **Number of monthly identified patients (monthly active: 3 million)** | **Number of total identified patients (total: 45 million)** |
|---|---|---|
| **DyKoNeM (Ours)** | **107,987** | **1,619,811** |
| Pedersen et al. (2019) | 74,716 | 1,120,744 |
| Ramachandran et al. (2020) | 71,717 | 1,075,756 |
| Pradier et al. (2020) | 80,419 | 1,206,292 |
| Maskew et al. (2022) | 66,744 | 1,001,164 |
| Bennemann et al. (2022) | 92,867 | 1,393,005 |
| Mütze et al. (2022) | 28,391 | 425,873 |
| Giesemann et al. (2023) | 79,968 | 1,199,532 |

**Table 18. Consultation Failure Reasons and Strategies for Intervention or Improvement**

<table>
<tr><th colspan="2">Failure Reason</th><th>Description</th><th colspan="2">Intervention or Improvement Strategies</th></tr>
<tr><td rowspan="3">Reasons attributed to the doctors</td><td>Ineffective or incomplete response</td><td>The medical consultation the user had received was ineffective, leaving the patient's inquiries not completely resolved (Pogorzelska et al., 2023).</td><td rowspan="3">Address through <i>early prediction</i></td><td>● Train or notify doctors to provide detailed explanations, medical advice, and additional information beyond patients' questions.<br>● Acquaint doctors with patient-doctor communication skills.</td></tr>
<tr><td>Long-time response delay</td><td>The extended amount of time a patient spends waiting before receiving the doctor's response after conveying their questions online (Yang et al., 2019).</td><td rowspan="2">● Employ mechanism to monitor doctors' response time.<br>● Recommend doctors that promptly reply to patients.<br>● Recommend some patient educating contents(i.e. articles, videos) to save patients' time cost during waiting.</td></tr>
<tr><td>No physician response</td><td>The doctor is not able to respond to the user in prescribed consultation time (Yan et al., 2020).</td></tr>
</table>

<table>
<tr><td rowspan="4"></td><td>Lack of operational advice</td><td>The doctor fails to provide explicit operational instruction (Yan et al., 2020).</td><td rowspan="4">Address through <i>accurate prediction</i> and analysis of failure reasons</td><td>Train doctors to provide operational instructions to ease patients' feeling of uncertainty.</td></tr>
<tr><td>Ambiguous answer or no explanation</td><td>The doctor's response is ambiguous (Yan et al., 2020).</td><td>● Train doctors to understand patients' knowledge limitations.<br>● Notify doctors to provide detailed explanations and medical information.</td></tr>
<tr><td>Lack of diagnosis information</td><td>Patient requirements are not satisfied when doctors refuse to provide a diagnosis based on limited information (Ahmad et al., 2006).</td><td>● Notify doctors to provide reasonable explanations when they refuse to provide diagnosis.<br>● Notify patients to provide adequate, accurate and relevant medical information for diagnosis (if possible).<br>● Acquaint patients with the limitations of online consultation.</td></tr>
<tr><td>Lack of emotional comfort</td><td>The doctors do not show enough empathy or emotional support during consultation (Yan et al., 2020).</td><td>● Notify doctors about patients' emotional needs.<br>● Acquaint doctors with effective communication that satisfies patients' emotional needs.</td></tr>
<tr><td rowspan="5">Reasons attributed to the patients</td><td>Wrong medical department or doctor</td><td>Patients select the doctors that do not meet their needs or are not suitable for their situation.</td><td rowspan="4">Address through <i>early prediction</i> and doctor recommendation systems</td><td>● Deploy effective triage service for patients.<br>● Provide patient educating contents (i.e. articles, videos) to enhance their health literacy.</td></tr>
<tr><td>Mismatch between patient expectation and doctor response</td><td>Patients expect doctors to provide high-quality answers when they pay higher consultation fees (Geng et al., 2024).</td><td>● Train doctors to provide both informational and emotional support to patients, detailed explanations, and additional information beyond patients' questions, especially those with high consultation fees.<br>● Acquaint patients with the limitations of online consultation.</td></tr>
<tr><td>Distrust toward the information</td><td>Patients doubt the correctness and reliability of doctor's advice (T. Lu et al., 2018; Yan et al., 2020; Yang et al., 2019).</td><td>● Notify doctors to provide detailed explanations for their diagnosis.<br>● Notify patients to clearly express their concerns and to elaborate on their historical treatment or consultations.</td></tr>
<tr><td>Decide not to consult</td><td>Patients do not want to consult anymore right after registration.</td><td>Provide effective triage service to patients.</td></tr>
<tr><td>Shift to offline consultation</td><td>Patients decide to shift to offline consultation with doctor's agreement (Y. Li et al., 2019).</td><td>-</td><td>Recommend the synchronous consultation service (i.e. phone-, video-based) to patients and doctors.</td></tr>
<tr><td rowspan="4">Reasons attributed to the platform or policy</td><td>Patient cannot buy target drugs</td><td>Doctors cannot prescribe drugs for patients on the platform due to limited authority or the drug is not available (Ding et al., 2020).</td><td>-</td><td>● Notify patients about doctors' diversified authentications in prescribing drugs.<br>● Provide a richer collection of drugs over the platform.</td></tr>
<tr><td>Platform shift</td><td>Patient agrees to shift to the doctor's affiliated hospital's online service platform.</td><td>-</td><td>Track impacts of platform shifting behavior and make relevant regulations for doctors.</td></tr>
<tr><td>Privacy concern</td><td>Patients are worried about the privacy of their consultation, and platforms failed to protect the historical record in patients' personal profiles (Z. Xu, 2019).</td><td rowspan="2">Address through accurate prediction and policy adjustments</td><td>● Acquaint patients with platform privacy protection policies to ease their anxiety.<br>● Enable patients to manage (i.e. hide, delete, add) their consultation records.</td></tr>
<tr><td>Inquiry quota limitation</td><td>Patient inquiries not resolved within the limited inquiry quota.</td><td>● Provide patients a flexible inquiry quota.<br>● Provide personalized service packages at their requests.</td></tr>
</table>

Concerning the implementation of our proposed method on the virtual health platform, we examine the effort required to apply this new approach, as well as the trade-offs between computational time and achieving improved results. Our research design concentrates on three types of knowledge: (1) Explicit knowledge: The real-time patient-doctor communication processes, which are readily accessible from the platform's IT database. Such knowledge can be represented by our framework and used in real-time predictions as it is available in the IT system. (2) Implicit knowledge: The digital traces of patients' interactions with various stakeholders during their healthcare journeys. We propose updating this knowledge at four-week intervals (see results in Table 9). The knowledge network updates can be performed offline during non-peak hours, representing a one-time cost over a long period, thus not impacting real-time predictions. (3) Knowledge fusion: For combining explicit and implicit knowledge, our proposed new fusion method emphasizes efficiency in algorithm design. Overall, compared to traditional methods for predicting patient consultation success, which rely solely on tabular patient information or patient-doctor dialogues (e.g., using LLM), our proposed method incurs slightly higher computational costs, some of which are one-time expenses. However, it offers significantly higher accuracy by providing personalized predictions based on implicit and explicit knowledge from each patient's healthcare journey and delivering timely predictions (see Figure 6).

Although promising, this study has several limitations. The first limitation arises from the use of data from a single virtual health platform. While the platform we collaborated with is among the largest in China and its data and profit model are highly representative, it is important to acknowledge the existence of diverse virtual health platforms that offer varied healthcare services. Additionally, healthcare systems vary across countries, implying that the development of virtual health and patient retention challenges may differ in each country. Consequently, it would be beneficial to expand our research to encompass multiple platforms and countries in order to conduct comparative studies in the future. Nevertheless, this hinges on future research to establish multilateral business partnerships. The second limitation relates to the issues of data security and privacy surrounding healthcare consultation success prediction in virtual health. The prediction process necessitates the integration and transmission of sensitive and confidential medical information across platforms, as well as online and offline healthcare systems, which can raise privacy and security concerns. Patients may be reluctant to share their personal and medical data online, especially if they have doubts about the security measures implemented by the virtual health platforms. In future research, numerous emerging information technologies can be employed to tackle this issue, including cutting-edge health information exchange technology, blockchain applications in healthcare, and more. Exploring these avenues will be crucial in advancing research on healthcare consultation success in virtual health settings. Next, we have included multiple stakeholders in the implicit knowledge network. In the future, intermediaries in healthcare systems, such as insurance and government entities, can also be incorporated into the implicit knowledge network. Moreover, our research focuses on text-based healthcare consultations. Consequently, explicit knowledge is derived primarily from textual information. Nonetheless, our methodology can be extrapolated to encompass audio and video-based consultations, utilizing such data as explicit knowledge to enhance the prediction of healthcare consultation success.

## CONCLUSION

Online healthcare consultations and virtual health, in general, have emerged as promising solutions to the global health burden and have attracted substantial investments. Consequently, this field has witnessed the emergence of numerous new platforms in various countries and regions, fostering intense competition where healthcare consultation success becomes paramount. We propose a novel framework that utilizes explicit and implicit knowledge suitable for predicting healthcare consultation success in the virtual context. Such methods can also be extended to other question domains with hybrid models incorporating both explicit detailed information and implicit digital traces, making them applicable in various problem domains.